\documentclass{article}

\usepackage{microtype}
\usepackage{graphicx}
\usepackage{booktabs} 

\usepackage{hyperref}

\usepackage[accepted]{icml2023}

\usepackage{amsmath}
\usepackage{amssymb}
\usepackage{mathtools}
\usepackage{amsthm}

\usepackage[capitalize,noabbrev]{cleveref}

\theoremstyle{plain}

\theoremstyle{definition}

\theoremstyle{remark}

\definecolor{goldenpoppy}{rgb}{0.99, 0.76, 0.0}
\usepackage[textsize=tiny]{todonotes}

\newcommand{\myparagraph}[1]{\vspace{0.1cm} \noindent {\bf #1}:}
\newcommand{\myparagraphb}[1]{\vspace{0.1cm} \noindent {\bf #1}}

\newcommand{\concat}{\mathbin\Vert}
\definecolor{darkyellow}{RGB}{209,208,30}
\usepackage{bbm}
\newcommand{\rulesep}{\unskip\ \vrule\ }
\usepackage{subcaption}

\icmltitlerunning{Hierarchical Neural Coding for Controllable CAD Model Generation}

\begin{document}

\twocolumn[
\icmltitle{Hierarchical Neural Coding for Controllable CAD Model Generation}



\icmlsetsymbol{equal}{*}

\begin{icmlauthorlist}
\icmlauthor{Xiang Xu}{sfu,ad,equal}
\icmlauthor{Pradeep Kumar Jayaraman}{ad}
\icmlauthor{Joseph G. Lambourne}{ad}
\icmlauthor{Karl D.D. Willis}{ad}
\icmlauthor{Yasutaka Furukawa}{sfu}
\end{icmlauthorlist}

\icmlaffiliation{sfu}{Simon Fraser University, Canada}
\icmlaffiliation{ad}{Autodesk Research}

\icmlcorrespondingauthor{Xiang Xu}{xuxiangx@sfu.ca}

\icmlkeywords{Computer Vision, Generative 3D Modeling, Computer Aided Design, Self-Supervised Learning, VQ-VAE, Controllable Generation, Transformer}

\vskip 0.3in
]



\printAffiliationsAndNotice{*Work partially done while interning at Autodesk.}  

\begin{abstract}
This paper presents a novel generative model for Computer Aided Design (CAD) that 1) represents high-level design concepts of a CAD model as a three-level hierarchical tree of neural codes, from global part arrangement down to local curve geometry; and 2) controls the generation or completion of CAD models by specifying the target design using a code tree. Concretely, a novel variant of a vector quantized VAE with ``masked skip connection'' extracts design variations as neural codebooks at three levels. Two-stage cascaded auto-regressive transformers learn to generate code trees from incomplete CAD models and then complete CAD models following the intended design. Extensive experiments demonstrate superior performance on conventional tasks such as unconditional generation while enabling novel interaction capabilities on conditional generation tasks. The code is available at \url{https://github.com/samxuxiang/hnc-cad}.
\end{abstract}

\section{Introduction}
\label{sec:intro}

From automobiles to airplanes, excavators to elevators, man-made objects are created using Computer Aided Design (CAD) software. Most modern CAD design tools employ the ``Sketch and Extrude" style workflow~\cite{camba2016parametric,Shahin2008}, where designers 1) draw loops of 2D curves as outer and inner boundaries to create 2D profiles; 2) extrude the 2D profiles to 3D shapes; and 3) add or subtract 3D shapes to build complex CAD models.

\begin{figure}[!t]
\includegraphics[width=0.99\columnwidth]{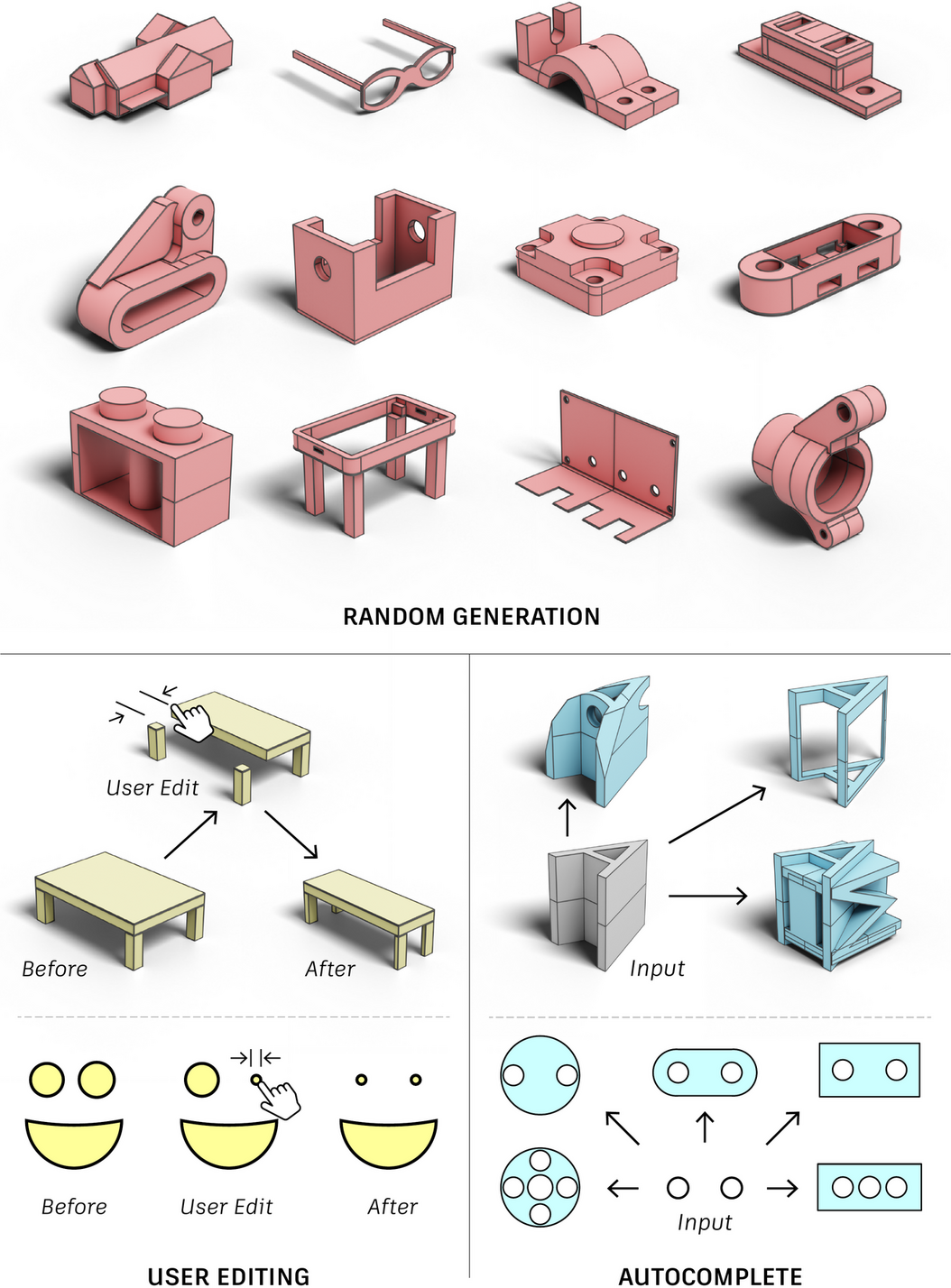}

  \caption{
  We propose three-level hierarchical neural coding for controllable CAD model generation. Our system learns high-level design concepts as discrete codes at different levels, enabling more diverse and higher-quality generation (top); novel user controls while specifying design intent (bottom-left); and autocompleting a partial CAD model under construction (bottom-right).
  }
  \vskip -0.1in
  \label{fig:teaser}
\end{figure}

CAD models created in this way have a natural tree structure which supports local edits. The curves at the leaves of the tree can be adjusted and the extrusions regenerated to update the final shape. For designers, it is also important that edits preserve ``design intent''.   Otey et al \cite{Otey2018} defines design intent as ``a CAD model’s anticipated behavior when altered'' while Martin \cite{Martin2023} describe it as ``relationships between objects, so that a change to one can propagate automatically to others''.  Although ``Sketch and Extrude'' allows local changes, it does not provide the relationships required to give the anticipated behavior when the model is edited. A computational system with understanding of design intent would revolutionize the practice of CAD. The system would help designers in 1) generating a diverse set of CAD models given high-level design concepts; 2) modifying existing CAD models while constraining certain model properties or 3) auto-completing designs interactively (See \autoref{fig:teaser}).

\begin{figure}[t]
    \begin{center}
        \includegraphics[width=0.99\columnwidth]{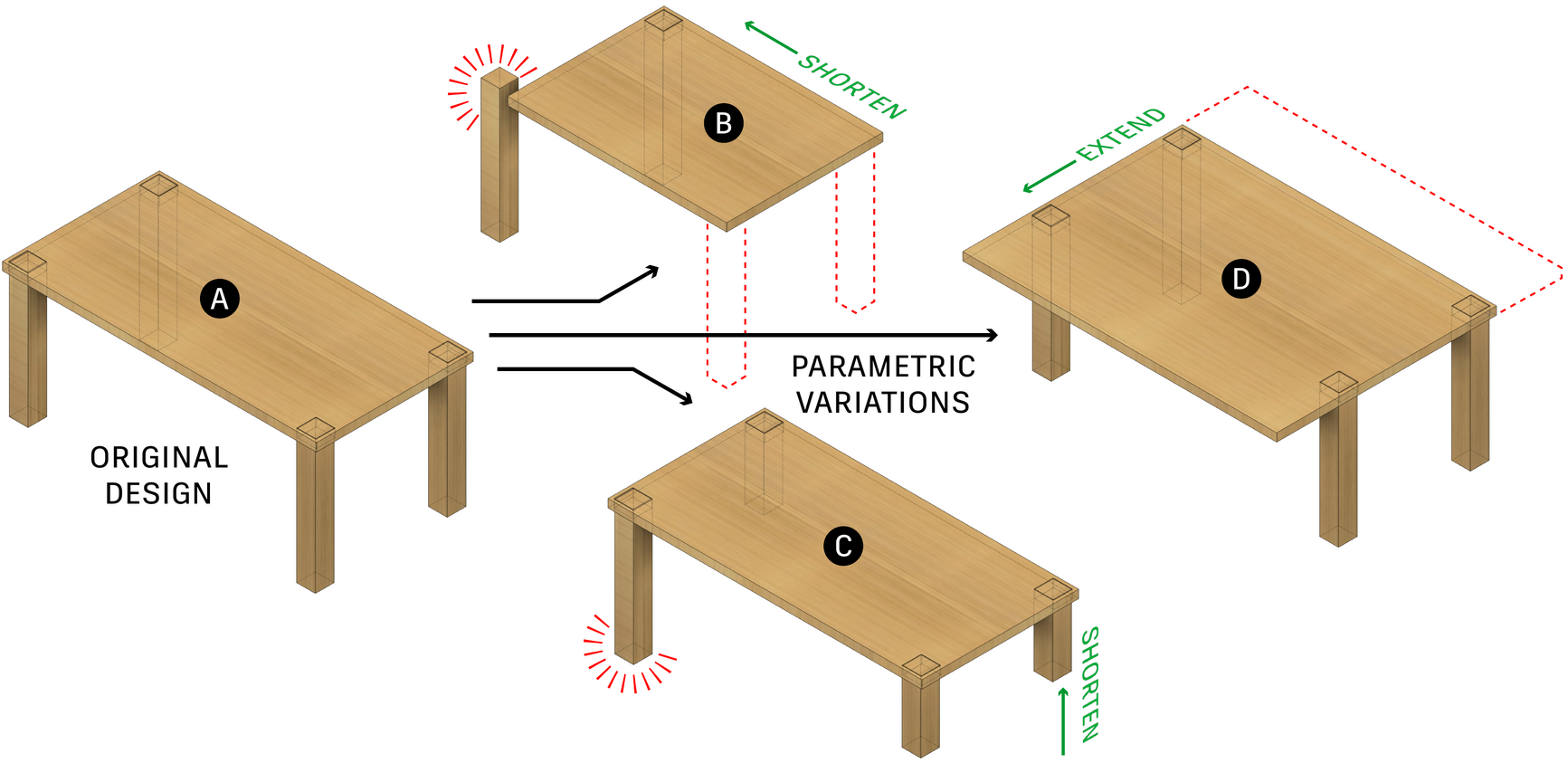}
        \caption{Example failures of parametric CAD, editing a design (a) by shortening or extending (green) the table. Inconsistent areas are highlighted in red.}
        \label{fig:problem}
    \end{center}
    \vskip -0.2in
\end{figure}

Unfortunately, such a system is not yet available for designers. A current industry standard is to manually specify parameters and equations which define the positions and sizes of profiles, and constraints to align geometry. This process, known as Parametric CAD, requires specialized skills~\cite{yares2013parametric} and easily breaks with unanticipated edits. \autoref{fig:problem} illustrates examples, where editing the geometry of a poorly constrained CAD model breaks the original design intent.
State-of-the-art research employs machine learning techniques to automatically generate CAD models, e.g. \citet{wu2021deepcad}. However, existing works do not make use of the hierarchical nature of CAD designs to provide effective design control.

This paper presents a novel generative network that captures the design intent of a CAD model as a three-level tree of neural codes, from local geometric features to global part arrangement; and controls the generation or completion of CAD models subject to the design intent specified by the code tree or an incomplete CAD model. CAD models are generated as sequences of modeling operations, then converted into the industry standard boundary representation (B-Rep) format for editing in mechanical CAD software.

Concretely, a novel variant of the vector quantized VAE~\cite{van2017neural} with ``masked skip connection'' learns design variations as three neural codebooks from a large-scale sketch-and-extrude CAD dataset~\cite{wu2021deepcad}. The masked skip connection is simple yet effective at extracting well-abstracted codebooks, making the relationships of codes and generated geometry intuitive.
Then, two-stage cascaded auto-regressive transformers learn to generate 1) three-level code trees given an incomplete CAD model, 2) complete CAD model given the code tree and the incomplete data. Designers can also directly provide a code tree for model generation.

Qualitative and quantitative evaluations against other generative baselines show that our system generates more realistic and complicated models in a random generation task. In user-controlled conditional generation tasks, our system demonstrates flexible and superior geometry control, enabled by the hierarchical code tree representation, over the current state-of-the-art deep learning-based generative models (i.e., SkexGen~\cite{xu2022skexgen}, DeepCAD~\cite{wu2021deepcad}). In summary, we make the following contributions:

\noindent \ $\bullet$
A neural code tree representation encoding hierarchical design concepts that enables generation of high quality and complex models, design intent aware user editing, and design auto-completion.

\noindent \ $\bullet$ A novel variant of VQ-VAE with a masked skip connection for enhanced codebook learning.

\noindent \ $\bullet$ State-of-the-art performance in CAD model generation over the previous SOTA methods.

\section{Related Work}
\label{sec:related_work}
\myparagraph{Constructive Solid Geometry (CSG)}
CSG builds complex shapes as Boolean combinations of simple primitives. Recent works utilized this representation for reconstructing CAD shapes with program synthesis~\cite{du2018inversecsg,nandi2017programming,nandi2018functional,sharma2018csgnet,ellis2019write,tian2019learning}, and unsupervised learning~\cite{kania2020ucsg,ren2021csg,chen2020bsp,yu2022capri,yu2023dualcsg}. Although a CSG tree can be converted into B-rep by building equivalent primitives and applying Boolean operations with solid modeling kernel, parametric CAD~\cite{camba2016parametric}, where a sequence of 2D sketches are built and extruded to 3D, is the dominant paradigm for designing mechanical parts and supports easy parametric editing.

\myparagraph{Direct CAD Generation}
Some recent works focused on directly generating CAD models without any supervision from CAD modeling sequences, by building the geometry of parametric curves~\cite{wang2020pie} and surfaces~\cite{sharma2020parsenet} with fixed~\cite{smirnov2021patches} or arbitrary topology for sketches~\cite{willis2021engineering} and solid models\cite{wang2022neuralface, guo2022complexgen,jayaraman2022solidgen}.
We focus more on controllable generation of parametric CAD in the form of sketch and extrude sequences.

\myparagraph{Sketch and Extrude CAD Generation}
Recent availability of large-scale datasets for parametric CAD has enabled learning based methods to leverage the CAD modeling sequence history~\cite{willis2020fusion,wu2021deepcad,xu2022skexgen} and sketch constraints~\cite{seff2020sketchgraphs} to generate engineering sketches and solid models.
The generated sequences can be parsed with a solid modeling kernel to obtain editable parametric CAD files containing 2D engineering sketches~\cite{willis2021engineering,para2021sketchgen,ganin2021computer,seff2021vitruvion} or 3D CAD shapes~\cite{wu2021deepcad,xu2022skexgen}.
Additionally, the generation can be influenced by a target B-rep~\cite{willis2020fusion,xu2021zone}, sketches~\cite{li2020sketch2cad,seff2021vitruvion}, images~\cite{ganin2021computer}, voxel grids~\cite{lambourne2022} or point clouds with~\cite{Point2CylUy} and without sequence guidance~\cite{ren2022extrude,li2023secad}. But this kind of control is on a global level, while we aim for hierarchical control on both global and local levels to support applications like design preserved edits and autocomplete.

\myparagraph{User-Controlled CAD Generation}
Providing user control over the generation process, while preserving design intent, is key for adoption of generative models in real world CAD software.
Although previous approaches can produce diverse shapes based on high level guidance, enabling user control over the generation process is more challenging. In the Sketch2CAD framework~\cite{li2020sketch2cad}, a network is trained to predict CAD operations that correspond with segmented sketch strokes, enabling a user interface for sketch based CAD modeling. 
Free2CAD~\cite{li2022free2cad} generalizes this system by additionally learning how to segment a complete sketch into groups that can be mapped to CAD operations. These works focus on localized control over the design process, and require significant user input. Recent works also leverage text prompts~\cite{wu2023iconshop,kodnongbua2023zero} and user-specified guidelines~\cite{cheng2023play}. SkexGen~\cite{xu2022skexgen} allows users to explore design variations with disentangled global control over over the topology and geometry of CAD shapes.
However, their approach simply aids in creating a new design from scratch and cannot be easily modified to provide an interactive experience that users expect for smartly editing CAD models or autocompleting their next steps to save effort.
Different from existing works, our method leverages the natural hierarchies which exist inside the CAD models to provide both global and local control over the generation process.

\section{Hierarchical CAD Properties}
\label{sec:cad_property}

\begin{figure}
    \begin{center}
        \includegraphics[width=0.99\columnwidth]{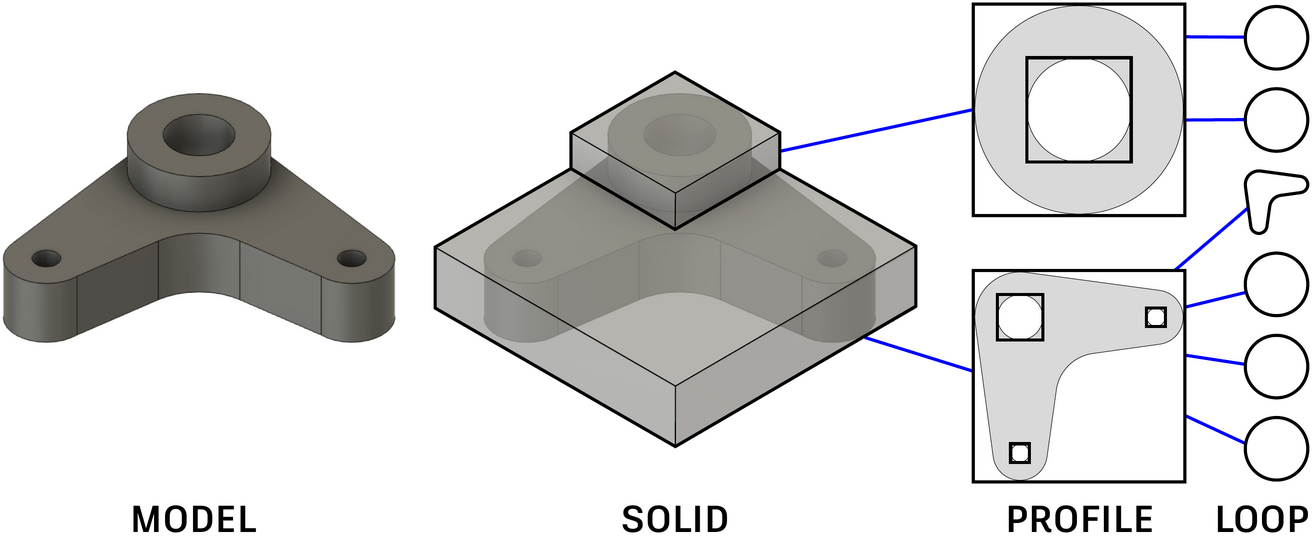  }
        \caption{Our hierarchical tree representation of a CAD model, with which a novel VQ-VAE learns codebooks at the levels of solid, profile, and loop.}
        \label{fig:representation}
    \end{center}
    \vskip -0.1in
\end{figure}

A sketch and extrude CAD model is naturally hierarchical (see \autoref{fig:representation}) with a \textit{loop} defining a closed path of connected curves, a \textit{profile} defining a closed area in the sketch plane bounded by one outer loop and some inner loops, and a \textit{solid} representing a set of extruded profiles that are combined to form the entire model.
Our goal is to enable local and global control in the generation of CAD models where users edit any of these entities and expect the rest to be updated sensibly automatically. To achieve this, we capture this hierarchy in the latent space of our neural networks. At higher levels of the hierarchy, the network learns the relative positions of lower level geometric entities, that is, the bounding boxes of the profiles and extrusions which make up the model. Concretely, we consider a CAD model as a (S)olid-(P)rofile-(L)oop tree:

\myparagraph{Loop ($L$)} At the leaf of the tree, we have loops. Each loop consists of a set of lines and arcs or a circle. 
The properties of a loop ($L$) is defined as a series of x-y coordinates separated by special \texttt{<SEP>} tokens:
\begin{equation}
L = \{(x_1, y_1), (x_2, y_2), {\small \texttt{<SEP>}}, (x_3, y_3),\ \ldots \}.
\label{eq:L}
\end{equation}
Lines are represented by the xy-coordinates of \textit{two} points. Here we use the start and end of the curve. Arcs are represented by \textit{three} points including start, middle and end point. Circles are represented by \textit{four} equally spaced points lying on the curve. With this representation, the curve types can be identified by the number of points as in~\cite{willis2021engineering}. 
We sort the curves in a loop so that the initial curve is the one with the smallest starting point coordinate, and the next one is its connected curve in counterclockwise order. 

\myparagraph{Profile ($P$)} The profile is above the leaf level. Since the loop geometry is captured at the leaf level, the properties of a profile node is defined as a series of 2D bounding box parameters of the loops within the sketch plane:
\begin{equation}
P = \{(x_i, y_i, w_i, h_i) \}_{i=1}^{N_{i}^{\text{loop}}}.
\label{eq:P}
\end{equation}
$i$ is the index of the $N_i^\text{loop}$ loops within a profile. $(x_i, y_i)$ is the bottom-left corner of the bounding box. $(w_i, h_i)$ is the width and height. We determine the order of bounding box parameters in profile $P$ by sorting the bottom-left corner of all the 2D bounding boxes in ascending order.

\myparagraph{Solid ($S$)}
Above the profile level, we have the 3D solid model formed by extruding one or more profiles.
The properties of a solid node captures the arrangement of extruded profiles using a series of 3D bounding box parameters:
\begin{equation}
S = \{(x_j, y_j, z_j, w_j, h_j, d_j)\}_{j=1}^{N_j^{\text{profile}}}.
\label{eq:S}
\end{equation}
$j$ is the index of the $N_j^\text{profile}$ extruded profiles within a model. $(x_j, y_j, z_j)$ is the bottom-left corner of the bounding box and $(w_j, h_j, d_j$) is its dimension. Likewise, the parameters in $S$ is sorted by the bottom-left corner of all the extruded 3D bounding boxes in ascending order.

\begin{figure*}
    \begin{center}
        \includegraphics[width=\textwidth]{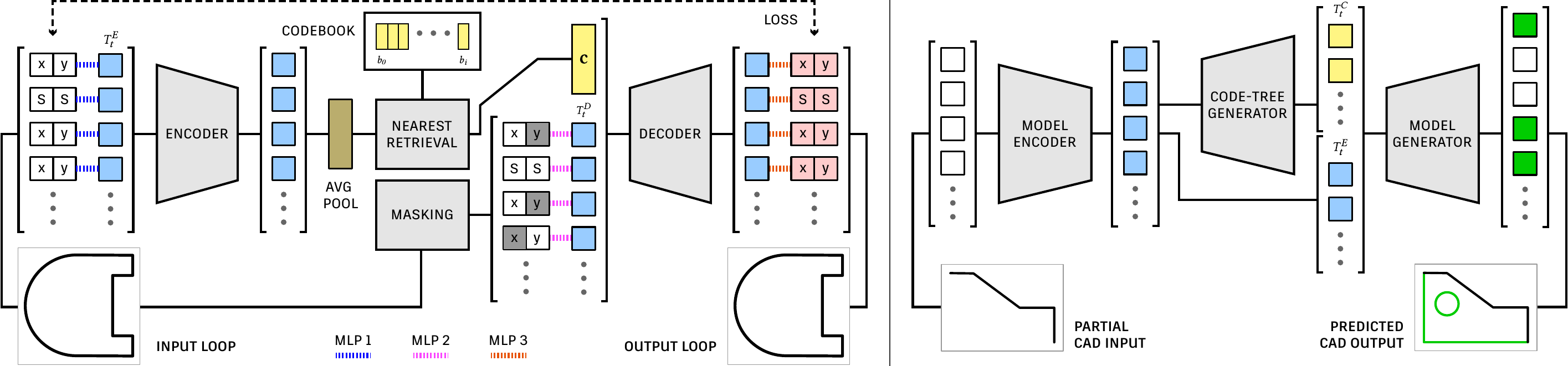}
        \caption{Left: VQ-VAE with masked skip connection for codebook learning. Given a CAD model as a construction sequence (e.g., x, y, S), an MLP and a Transformer encoder convert the input to latent codes ({\color{blue}{ $T^E_t$}}), and a vector quantization extracts a code ({\color{goldenpoppy}{$\mathbf{c}$}}) after average pooling.
        A Transformer decoder recovers the input sequence, conditioned on the vector-quantized code ({\color{goldenpoppy}{$\mathbf{c}$}}) and the masked input sequence ({\color{blue}{ $T^D_t$}}). Grey color represents input tokens that were masked out. 
        Right: Controllable CAD generation module with two-stage auto-regressive generators. Given a partial CAD model, a model encoder converts it to latent embeddings ({\color{blue}{$T^E_t$}}). The first auto-regressive Transformer generates hierarchical neural codes ({\color{goldenpoppy}{$T^C_t$}}) conditioned on the encoded embeddings. The second auto-regressive Transformer generates a new CAD model.}
        \label{fig:architecture}
    \end{center}
    \vskip -0.1in
\end{figure*}

\section{Three-Level Codebook Learning}
\label{sec:codebook_learning}

Given a dataset of sketch and extrude CAD models in the (S)olid-(P)rofile-(L)oop tree format, a novel variant of the vector quantized VAE (VQ-VAE)~\cite{van2017neural,razavi2019generating} learns their latent  patterns as three discrete codebooks, which encode a CAD model into a tree of neural codes for downstream applications.

Following SkexGen \cite{xu2022skexgen}, the foundation of our architecture for learning codebooks is a VQ-VAE, consisting of a Transformer encoder $E$ and decoder $D$ (see \autoref{fig:architecture}). We learn (L)oop, (P)rofile, and (S)olid codebooks independently. Different from SkexGen and previous work on masked learning~\cite{he2022masked}, we apply masking on a skip-connection from the encoder input to the decoder input. Intuitively, a standard VQ-VAE (i.e., without skip connection) is trained to recover instance-specific input details, which would be a challenge for the quantized code if it is learning instance-agnostic design patterns. A na\"ive skip connection allows the decoder to cheat by directly copying the input. Masking the skip connection forces the decoder to relate partial details from unmasked elements and fill-in missing ones, where the relation is guided by design patterns encoded in the code. 

\myparagraph{Encoder}
Consider a (L)oop node $L$ (\autoref{eq:L}), containing a series of x-y coordinates and special \texttt{<SEP>} tokens. We use a $65$D one-hot vector to represent a token, where a coordinate is quantized to a 6 bit (i.e., 64D)~\cite{xu2022skexgen, seff2021vitruvion} and \texttt{<SEP>} requires one extra dimension. Let $T^E_t$ denote the 256D embedding of the $t^{\text{th}}$ token for the Transformer encoder. The embedding is initialized as:
\begin{eqnarray}
T^E_t \leftarrow
\begin{cases}
  \text{MLP}(W_\text{emb}x_t \concat W_\text{emb}y_t) + \gamma_t \ \ \text{ (for x-y)},\\
  \text{MLP}(W_\text{emb}\texttt{<SEP>} \concat W_\text{emb}\texttt{<SEP>}) + \gamma_t .\\
\end{cases}
\label{eq:mlp_enc}
\end{eqnarray}
$W_{\text{emb}}$ is a $65\times32$ token embedding matrix. $\concat$ is the concatenation operator. MLP is a 2-layer multilayer perception. $\gamma_t$ is a learnable 256D positional embedding. Second case is for \texttt{<SEP>} where value is repeated twice. For (P)rofile and (S)olid codebooks, we process each of the 2D or 3D bounding box parameters the same way as $x_t$, $y_t$ coordinates, except with no \texttt{<SEP>} tokens.

\myparagraph{Vector Quantization}
The outputs of the encoder ($E$), with sequence length $T$, are first average pooled, forming $\overline{E}(T^E)$. The standard vector quantization procedure is then applied to obtain a 256D codebook vector $\mathbf{c}$. More specifically, we compare the Euclidean distance between codebook vector $\mathbf{b}$ and encoded $\overline{E}(T^E)$ and perform a nearest neighbor lookup.
\begin{eqnarray}
\mathbf{c} \leftarrow \mathbf{b}_k, \quad \text{where} \hspace{0.1cm} k = \mbox{argmin}_i \left|\left| \overline{E}(T^E) - \mathbf{b}_{i} \right|\right|^2.
\label{eq:code_quan}
\end{eqnarray} 

\myparagraph{Decoder with Masked Skip Connection}
The decoder takes the quantized code $\mathbf{c}$ and the input series of x-y coordinates and \texttt{<SEP>} tokens with masking, and predicts the masked tokens. For example, in the case of a loop node, any of the $x_t$, $y_t$ and \texttt{<SEP>} tokens could be masked (concretely 30\% to 70\% of the tokens per model randomly). Let $T^D_t$ denote the embedding of the $t_{\text{th}}$ token as an input to the decoder.
Each token is embedded exactly as in \autoref{eq:mlp_enc}, except that embeddings of masked tokens are replaced with a learnable shared 32D mask token embedding $m$. 

The 256D codebook vector $\mathbf{c}$ from the encoder is concatenated together with $\{T^D_t\}$ and passed to the decoder ($D$), which has 4 self-attention layers. The idea here is to force the encoder to learn useful latent features that can help the decoder to predict the masked tokens. Finally, an MLP is applied to each token embedding (except the codebook vector) after the decoder to produce $(2 \times 65)$D logits, a pair of probability values over the 65 class labels for predicting the xy-coordinates or the \texttt{<SEP>} token.

\myparagraph{Loss Function} The training loss consists of three terms:
\begin{eqnarray}
\label{eq:mae_loss_fn}
&&\sum_t \text{EMD}\Big(D(\mathbf{c}, \{ T^D_t\}) , \text{ } \mathbbm{1}_{T_t} \Big) + \nonumber \\
&&\left| \left| sg[\overline{E}(T^E)] - \mathbf{c} \right| \right|^2_2 + \beta \left| \left| \overline{E}(T^E) - sg[\mathbf{c}] \right| \right|^2_2. \label{eq:codebook_loss}
\end{eqnarray} 
The first term is the squared Earth Mover’s Distance Loss between the decoder output probability and the corresponding data property's one hot encoding $\mathbbm{1}_{T_t}$. The loss is only applied at masked tokens. We use the EMD loss function from~\cite{hou2016squared} which assumes ordinal class labels and penalizes predictions closer to the ground-truth less than those far away. This works better than a cross-entropy loss since x-y coordinates carry distance relations, allowing the loss to focus on predictions far away from the ground-truth.  Note that we treat the \texttt{<SEP>} token in loop data properties differently by applying the standard cross-entropy loss on it as this is not an ordinal class label.

The second and third terms are the codebook and commitment losses used in VQ-VAE~\cite{van2017neural,razavi2019generating}. $sg$ denotes the stop-gradient operation, which is the identity
function in forward pass but blocks gradients in backward pass. $\beta$ scales the commitment loss and is set to 0.25. We use the exponential moving average updates with a decay rate of 0.99~\cite{razavi2019generating}.

\section{Controllable CAD Generation}
\label{sec:control_gen}

Loop, profile, and solid codebooks allow us to express the design concepts of a CAD model as hierarchical neural codes, enabling diverse and high-quality generation, novel user controls specifying design intent, and autocompletion of incomplete CAD models. Concretely, given an incomplete CAD model as a sketch and extrude construction sequence: 1) A model encoder turns the input sequence into latent embeddings; 2) An auto-regressive Transformer generates a code tree, conditioned on the embedded input sequence; and 3) The second auto-regressive Transformer generates the full CAD models, conditioned on the embedded input sequence and a code tree.

\myparagraphb{Model Encoder:} The model encoder backbone is the standard Transformer encoder module with 6 self-attention layers. We borrow the format used in SkexGen~\cite{xu2022skexgen} and represent a model as a sequence of tokens, each of which is a one-hot vector, uniquely determining a curve type, quantized curve parameter and quantized extrusion parameter. The encoder converts the one-hot vectors into a series of 256D latent embeddings $\{T^E_t\}$.~\footnote{As in SkexGen, we encode ``geometry'' and ``extrusion'' sequence separately and concatenate the embeddings to get $T^E_t$. 
For experiments with 2D sketches, only the geometry encoder is used.}

\myparagraphb{Code Tree Generator:} $G_{\text{code}}$ is an autoregressive decoder which generates a hierarchy of codes $\{T^C_t\}$.   A code is assigned to each (S)olid, (P)rofile, or (L)oop from a corresponding codebook, conditioned on the encoded embeddings $\{T^E_t\}$. Similar to the hierarchical property representation (\autoref{sec:cad_property}), hierarchical codes are represented as a series of feature vectors indicating either a code or a separator token. Concretely, a feature is a one-hot vector whose size is the total number of codes in the three codebooks plus one for the separator.
For example, consider the code tree in \autoref{fig:representation}, consisting of a model with one solid, two profiles, and two or four loops. This tree is represented as features in the following order [S, $<$SEP$>$, P, L, L, $<$SEP$>$, P, L, L, L, L]. Here we perform depth-first traversal of the neural code tree and the boundary command $<$SEP$>$ is used to indicate a new grouping of profile and loop codes.

$G_{\text{code}}$ has 6 self-attention (SA) layers interleaved with 6 cross-attention (CA) layers. The first SA layer is over the query tokens $\{T^{\tilde{C}}_t\}$, each of which is initialized by a position encoding $\gamma_t$ and autoregressively estimated. The input to each of the CA layers is $\{T^E_t\}$. Each SA or CA layer has 8-heads attentions, followed by an Add-Norm layer.
A query token $\{T^{\tilde{C}}_t\}$ will have a generated code index, which is converted to a code $T^C_t$. A separator is replaced by a learnable embedding.
\begin{eqnarray}
T^C_t \leftarrow
\begin{cases}
  \text{Codebook}(T^{\tilde{C}}_t) + \gamma_t & \text{(for code)},\\
  W_\text{emb}\texttt{<SEP>} \hspace{0.44cm} + \gamma_t & \text{(for \texttt{<SEP>})}.
\end{cases}
\label{eq:first_ar_decoder}
\end{eqnarray}
$\text{Codebook}$ denotes the mapping from a code index to the code. We train $G_{\text{code}}$ with the standard cross-entropy loss. Note that for unconditional generation, we remove the partial CAD model encoder and train SA layers with query tokens ($\{T^{\tilde{C}}_t\}$) only, without cross-attention layers and $\{T^E_t\}$.

\myparagraphb{Model Generator:} The model generator is the second auto-regressive decoder $G_{\text{cad}}$, generating a sketch-and-extrude CAD model. $G_{\text{cad}}$ is the same as the SkexGen decoder~\cite{xu2022skexgen} except that partial CAD model embeddings $\{T^E_t\}$ and the hierarchical neural codes $\{T^C_t\}$ control the generation via the cross-attention layers, while SkexGen only allows the specification of global codes.
The architecture specification is the same as the first 
decoder. The query tokens ($T^{out}_t$) contain the generated CAD command sequences as one-hot vectors~\cite{xu2022skexgen}, where we use the same standard cross entropy loss.
\section{Evaluation}
\label{sec:experiments}

\begin{figure*}
    \centering
    \begin{subfigure}[t]{0.31\textwidth}
    \centering
    \includegraphics[width=\textwidth]{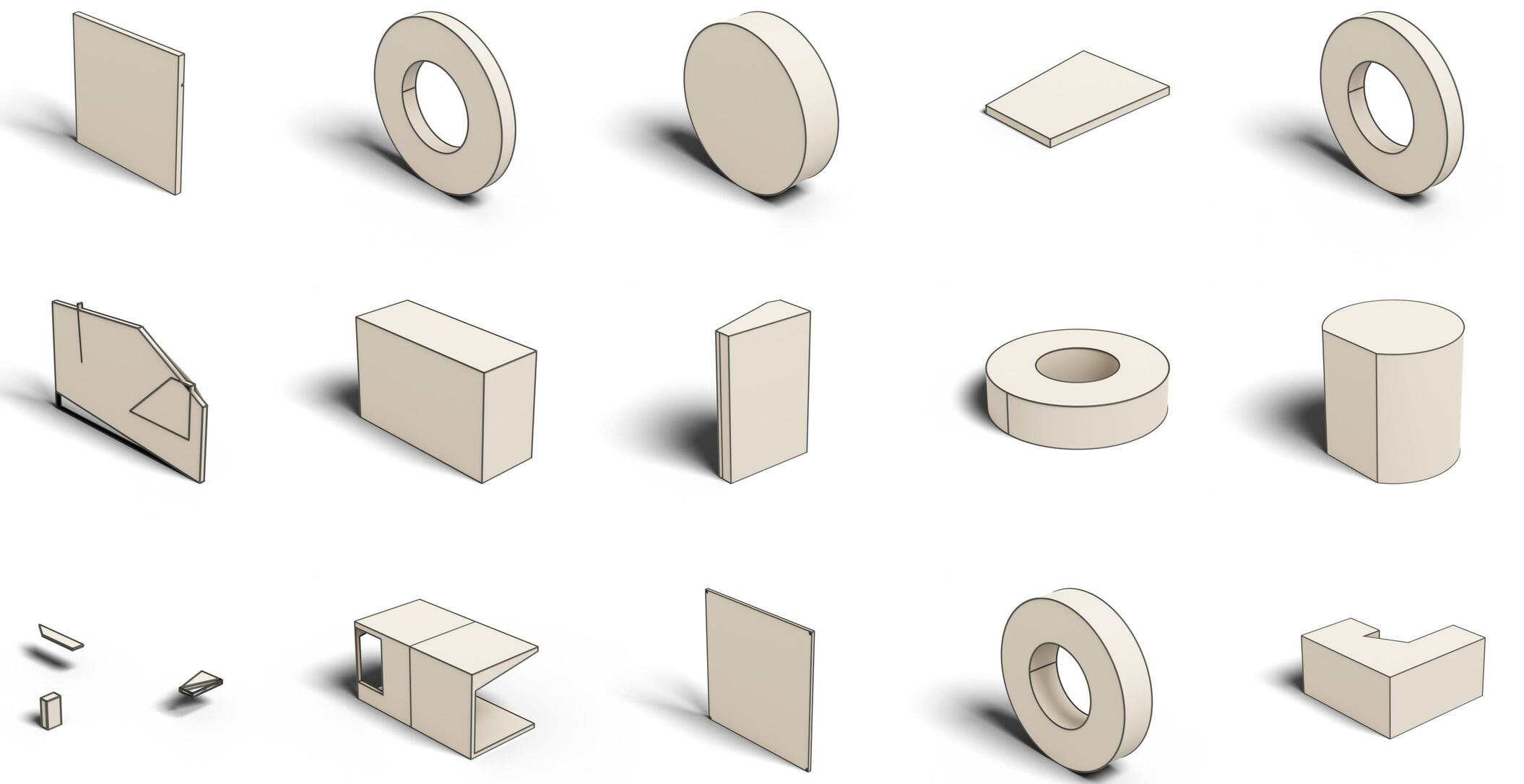}
    \end{subfigure}
    \rulesep
    \begin{subfigure}[t]{0.31\textwidth}
    \centering
    \includegraphics[width=\textwidth]{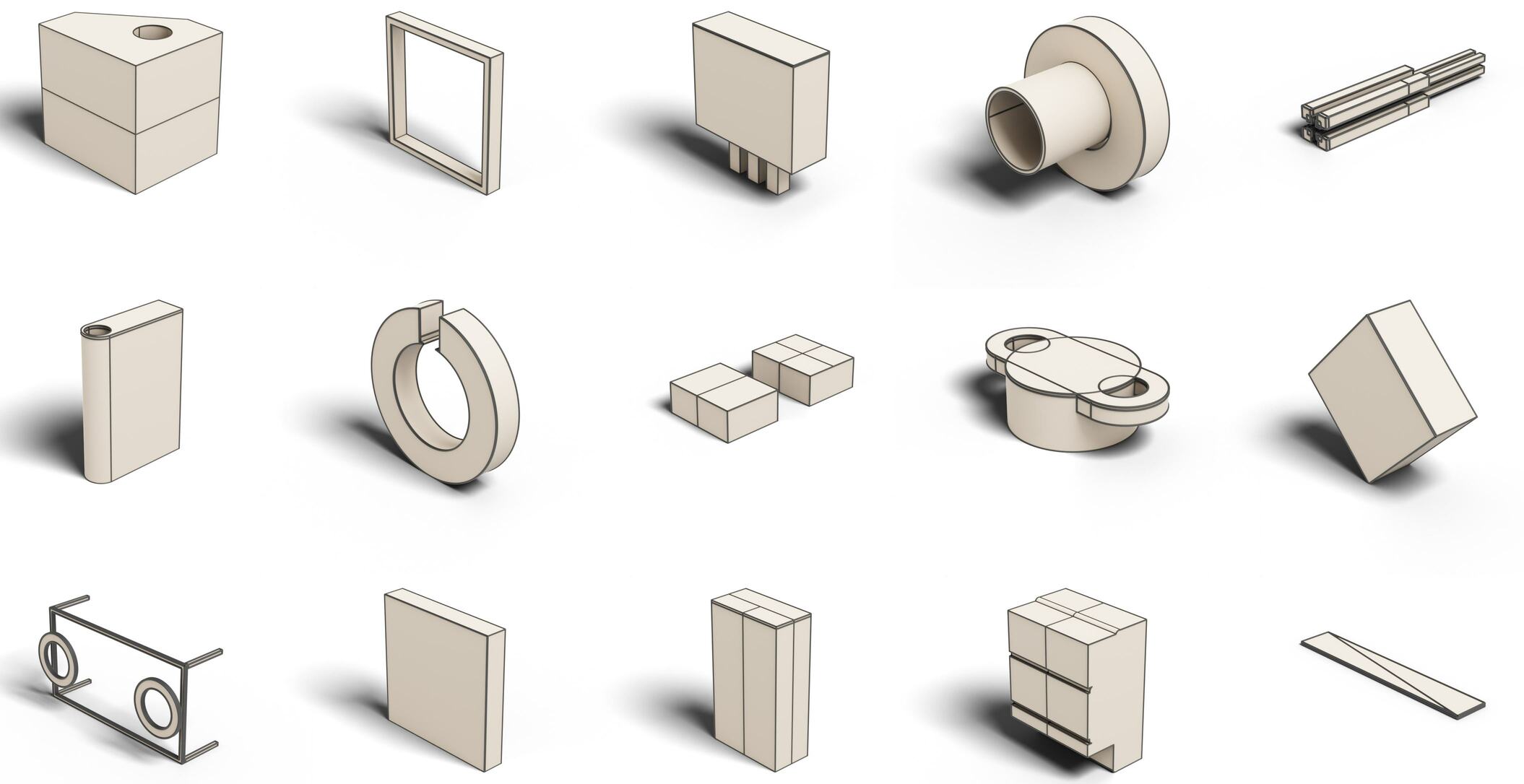}
    \label{fig:solid_uncond_skexgen_random}
    \end{subfigure}
    \rulesep
    \begin{subfigure}[t]{0.31\textwidth}
    \centering
    \includegraphics[width=\textwidth]{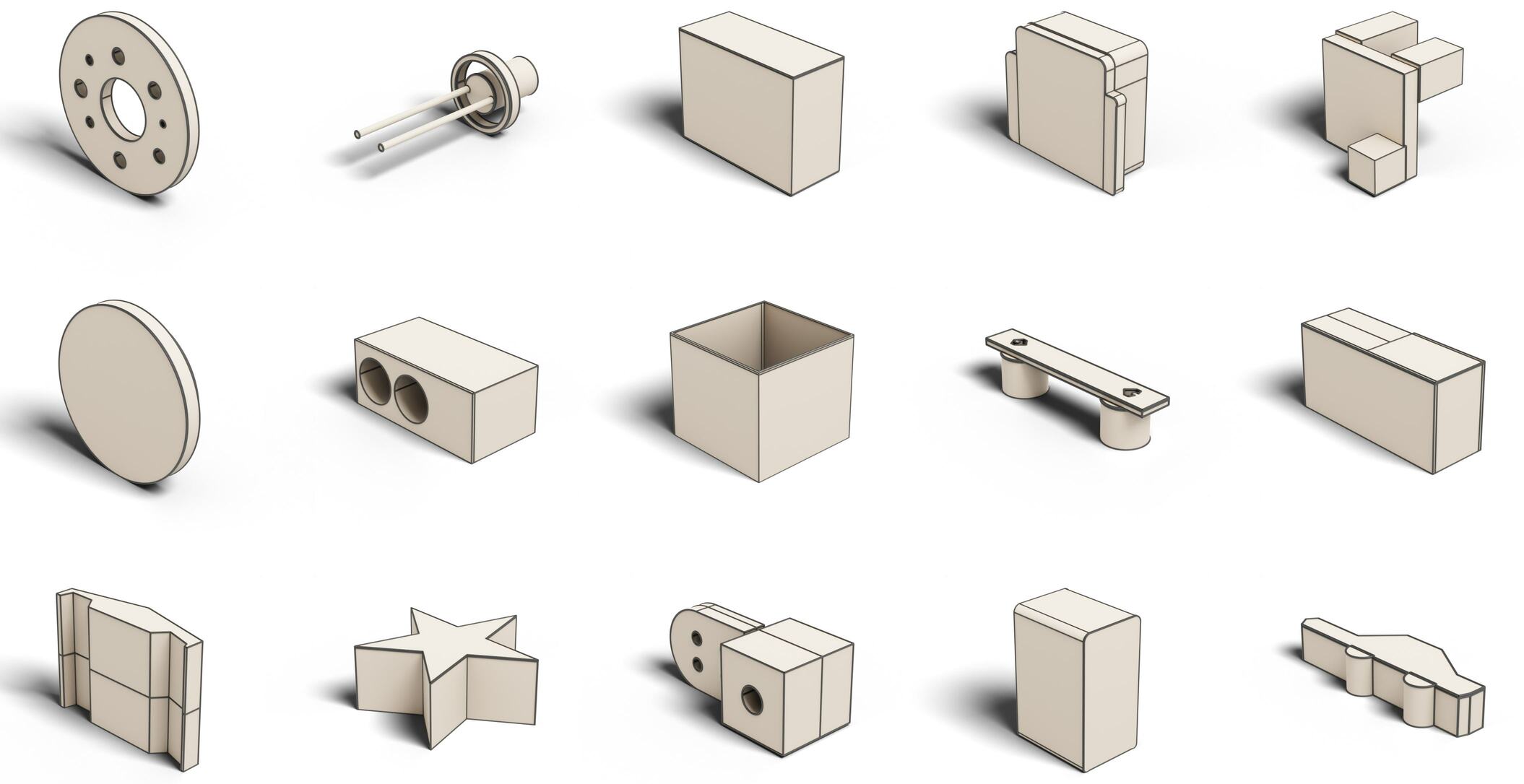}
    \end{subfigure}\\
    \begin{subfigure}[t]{0.31\textwidth}
    \centering
    \includegraphics[width=\textwidth]{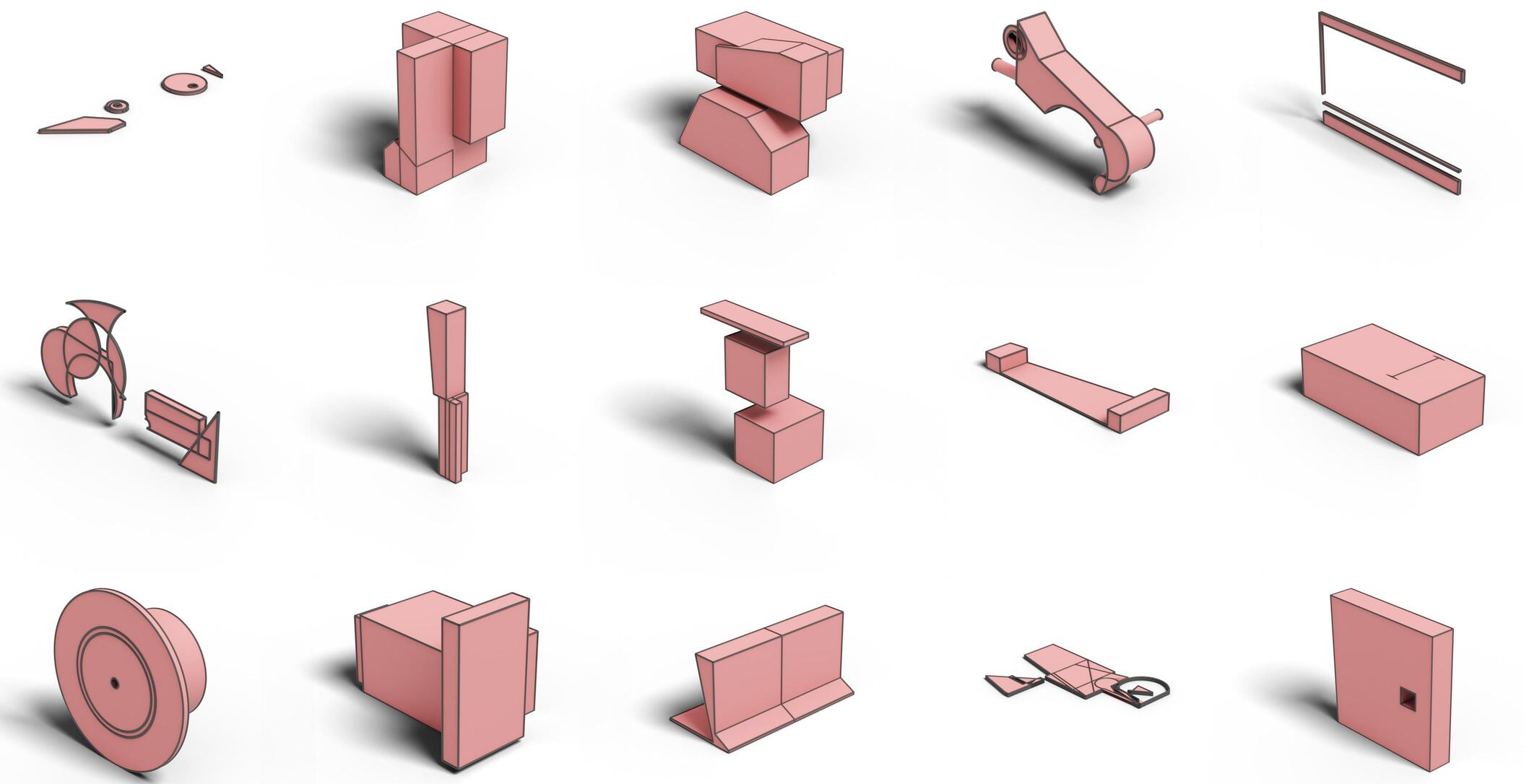}
    \caption{DeepCAD}
    \label{fig:solid_uncond_deepcad_complex}
    \end{subfigure}
    \rulesep
    \begin{subfigure}[t]{0.31\textwidth}
    \centering
    \includegraphics[width=\textwidth]{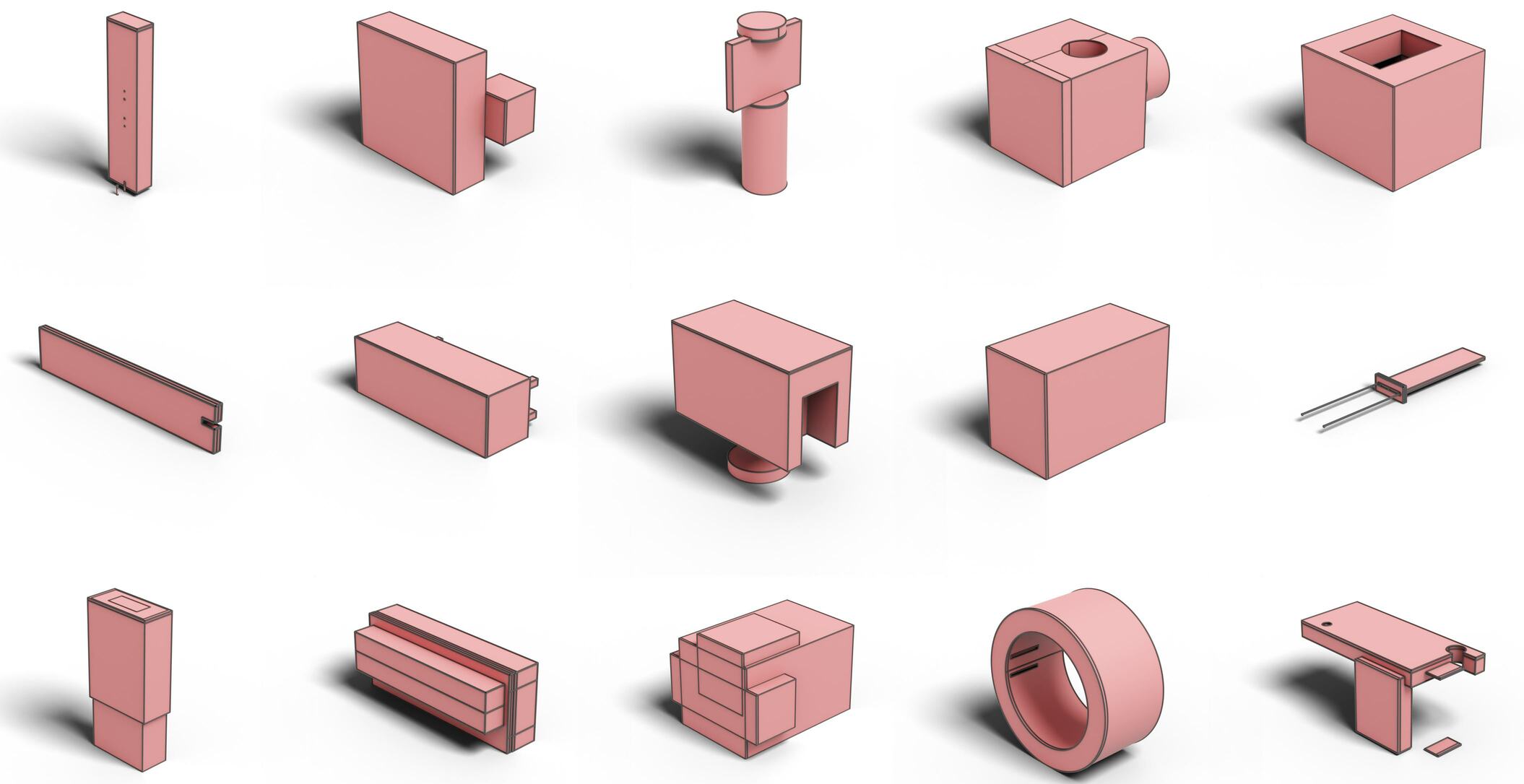}
    \caption{SkexGen}
    \label{fig:solid_uncond_skexgen_complex}
    \end{subfigure}
    \rulesep
    \begin{subfigure}[t]{0.31\textwidth}
    \centering
    \includegraphics[width=\textwidth]{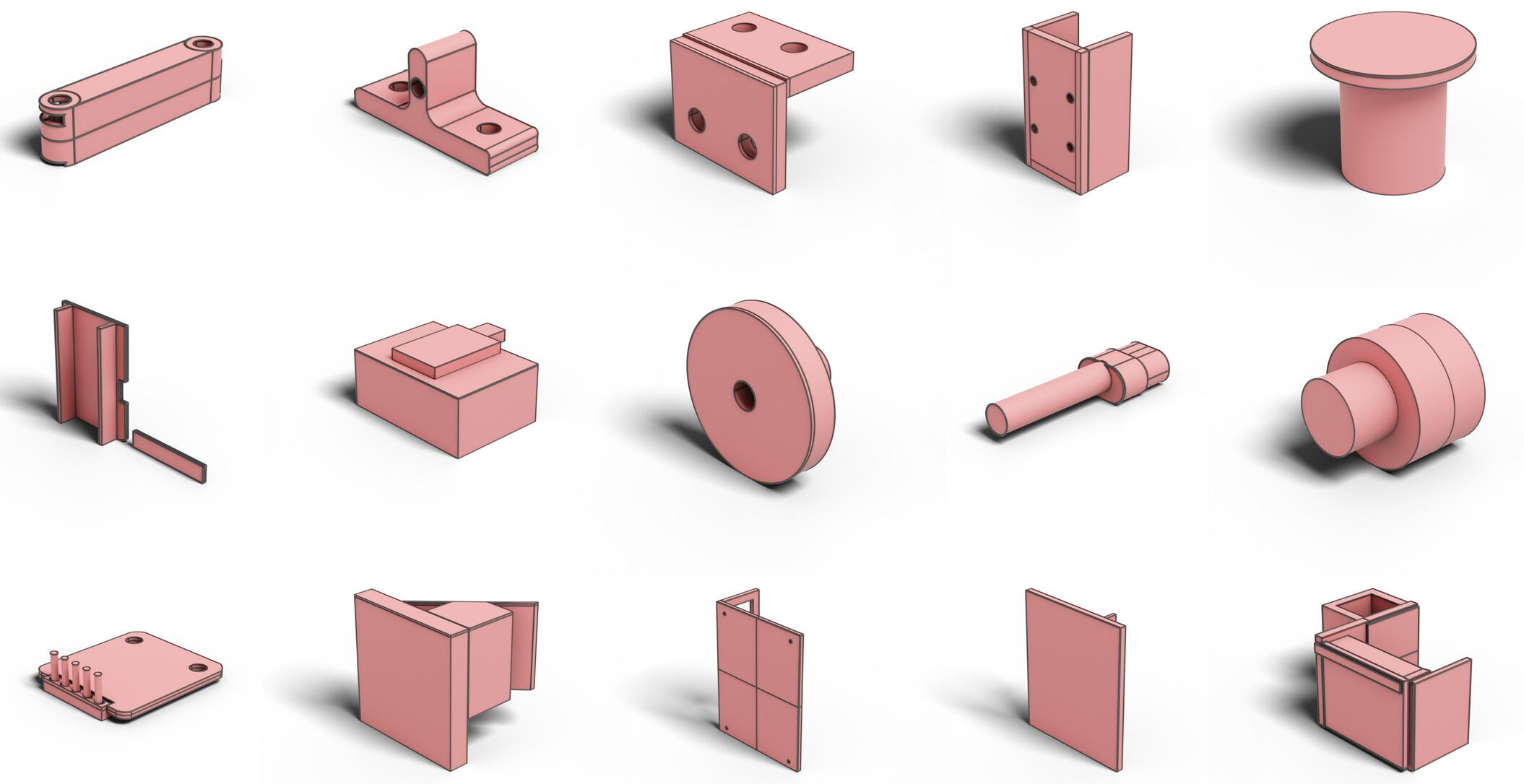}
    \caption{Ours}
    \label{fig:solid_uncond_ours_complex}
    \end{subfigure}
    \caption{Unconditional generation results by (a)  DeepCAD, (b) SkexGen and (c) our method. 
    The bottom three rows (red color) show complex samples
    with three or more sketch-extrude steps.}
    \label{fig:solid_uncond}
\end{figure*}

This section presents unconditional and conditional generation results, which demonstrate 1) Higher quality, diversity, and complexity compared to current state-of-the-art; 2) Controllable generation via hierarchical neural codes; and 3) Two important applications, user-edit and auto-completion.

\subsection{Experiment Setup}

\myparagraph{Dataset} We use the large-scale DeepCAD dataset~\cite{wu2021deepcad} with ground-truth sketch-and-extrude models.
DeepCAD contains 178,238 sketch-and-extrude models with a split of $90\%$ train, $5\%$ validation, and $5\%$ test samples. We detect and remove duplicate models from the training set as in prior works~\cite{willis2021engineering,xu2022skexgen}.
After extracting the hierarchical properties for (L)oop, (P)rofile, and (S)olid (\autoref{sec:cad_property}), we also remove duplicate properties for each level. Lastly, we use a CAD model for training only when the number of solids is at most 5, the number of loops is at most 20 for every profile, the number of curves is at most 60 for every loop, and the total number of commands in the sketch-and-extrude sequence is at most 200. After the duplicate removal and filtering, the training set contains 102,114 solids, 60,584 profiles, 150,158 loops for codebook learning, and 124,451 sketch-and-extrude sequences for CAD model generation training. For CAD engineering drawings, we follow SkexGen~\cite{xu2022skexgen} and extract sketches from DeepCAD. A total of  99,650 sketches are used for training after duplicate removal. 

\myparagraph{Implementation Details} Models are trained on an Nvidia RTX A6000 GPU with a batch size of 256. The codebook module and the generation module are trained for 250 and 350 epochs, respectively. We use the improved Transformer backbone with pre-layer normalization as in~\cite{wu2021deepcad,xu2022skexgen}. Input embedding dimension is $256$. Feed-forward dimension is $512$. Dropout rate is $0.1$. Each Transformer network in the generation module has 6 layers with 8 attention heads. The codebook learning networks have 4 layers. We use the AdamW~\cite{loshchilov2018decoupled} optimizer with a learning rate of $0.001$ after linear warm-up for 2000 steps.  At test time, we use nucleus sampling~\cite{holtzman2019curious} to autoregressively generate the codes and CAD tokens. To reduce overfitting, we follow~\cite{xu2022skexgen} and augment the training data by adding a small random noise to the input curve coordinates.

VQ-VAEs suffer from codebook collapse and we employ an approach from Jukebox~\cite{dhariwal2020jukebox} that reinitializes under-utilized codes (less than 7 mapped samples). To identify the optimal codebook size, we trained our model using different codebook sizes and evaluated the unconditional generation results using the $5\%$ validation set in DeepCAD. Our analysis revealed that the model performance was best for codebook size ranging from 2,000 to 4,000, with larger codebook not providing noticeable improvement. 
Our final codebook size is around 3,500 for profile and solid, and 2,500 for loop. The compression ratio of dividing the number of unique data by the codebook size is approximately 60x for loop, 17x for profile, and 29x for solid.

\subsection{Metrics}
Five established metrics quantitatively evaluate random generation. Three metrics are based on point clouds sampled on the model surfaces. Two metrics are based on generated tokens of sketch and extrude construction sequence.

\myparagraphb{Point-cloud} metrics measure generation diversity and quality by sampling 2,000 points on each generated or ground-truth data and compare two sets of point clouds~\cite{achlioptas2018learning,wu2021deepcad,xu2022skexgen}.

\noindent $\bullet$ \textit{Coverage} (COV) is the percentage of ground-truth models that have at least one matched generated sample. The matching process assigns every generated sample to its closest neighbor in the ground-truth set based on Chamfer Distance (CD) or Earth Mover's Distance (EMD).
COV measures the diversity of generated shapes. If CAD generation suffers from mode collapse, generated shapes would only match a few ground-truth models, leading to low coverage scores.

\noindent $\bullet$ \textit{Minimum Matching Distance} (MMD) reports the average minimum matching distance between the ground-truth set and the generated set. 

\noindent $\bullet$ \textit{Jensen-Shannon Divergence} (JSD) is the similarity between two probability distributions, measuring how often the ground-truth points clouds occupied similar locations as the generated point clouds. We voxelize the 3D space and count the number of points in each voxel. This gives us occupancy distributions for computing the JSD score. 

\myparagraphb{Token} metrics measure uniqueness~\cite{willis2021engineering}. Numeric fields are quantized to 6-bit.

\noindent $\bullet$ \textit{Novel} is the percentage of generated CAD sequence that does not appear in the training set. 

\noindent $\bullet$ \textit{Unique} is the percentage of generated data that appears once in the generated set. 

\begin{table}
\caption{Quantitative evaluations on the CAD generation task based on the \textit{Coverage} (COV) percentage, \textit{Minimum Matching Distance} (MMD),  \textit{Jensen-Shannon Divergence} (JSD), the percentage of \textit{Unique} and \textit{Novel} scores and \textit{Realism} as perceived by human evaluators. }
\label{tab:solid_uncond}
\begin{center}
\setlength\tabcolsep{3.5 pt}
\small
\begin{tabular}{lccccc|c}
\toprule
       Method  & COV   & MMD  & JSD  & Novel   & Unique & Realism    \\
         &  \% $\uparrow$  & $\downarrow$  & $\downarrow$  &\% $\uparrow$   & \% $\uparrow$  & \% $\uparrow$     \\ \midrule     
DeepCAD          &    80.62    &  1.10    &  3.29     &  91.7 & 85.8 & 38.7\\
SkexGen  &   84.74     &   1.02   &   0.90  & 99.1  & 99.8 & 46.9\\
Ours     &   87.73     &  0.96       &   0.68    &  93.9 & 99.7 & 49.2\\
\bottomrule
\end{tabular}
\end{center}
\end{table}

\subsection{Unconditional Generation}
We compare with two sketch-and-extrude baselines, DeepCAD~\cite{wu2021deepcad} and SkexGen~\cite{xu2022skexgen}, for the unconditional generation task. Our cascaded auto-regressive system generates a code tree and then a CAD model.
Each method generates 10,000 CAD models, which are compared with randomly selected 2,500 ground truth models from the test set.

\myparagraph{Quantitative Evaluation}
\autoref{tab:solid_uncond} reports the average scores across 3 different runs. Our method outperforms baselines on all three point cloud evaluation metrics, demonstrating great improvements in quality and diversity. The \textit{Unique} score of our method matches SkexGen and is significantly better than DeepCAD. For the \textit{Novel} score, our method is slightly worse than SkexGen, while still significantly better than DeepCAD, which is caused by the smaller training set that lacks diversity and has only a few complex shapes. SkexGen suffers less from this issue since it fails to generate very complex CAD models (see \autoref{fig:solid_uncond}).

\myparagraph{Qualitative Evaluation} 
\autoref{fig:solid_uncond} provides side-by-side qualitative comparisons at different steps of sketch-and-extrude. The figure shows that our approach generates well-structured CAD models, reminiscent of real-world examples. Generated solids have more complicated shape geometries and part arrangements. Additional qualitative results are available in \autoref{fig:mosaic_appendix} and \autoref{fig:complex_mosaic_appendix}. Also see \autoref{sec:appendix_sketch_results} for the sketch generation results.  

\begin{figure}
    \centering
\includegraphics[width=\columnwidth]{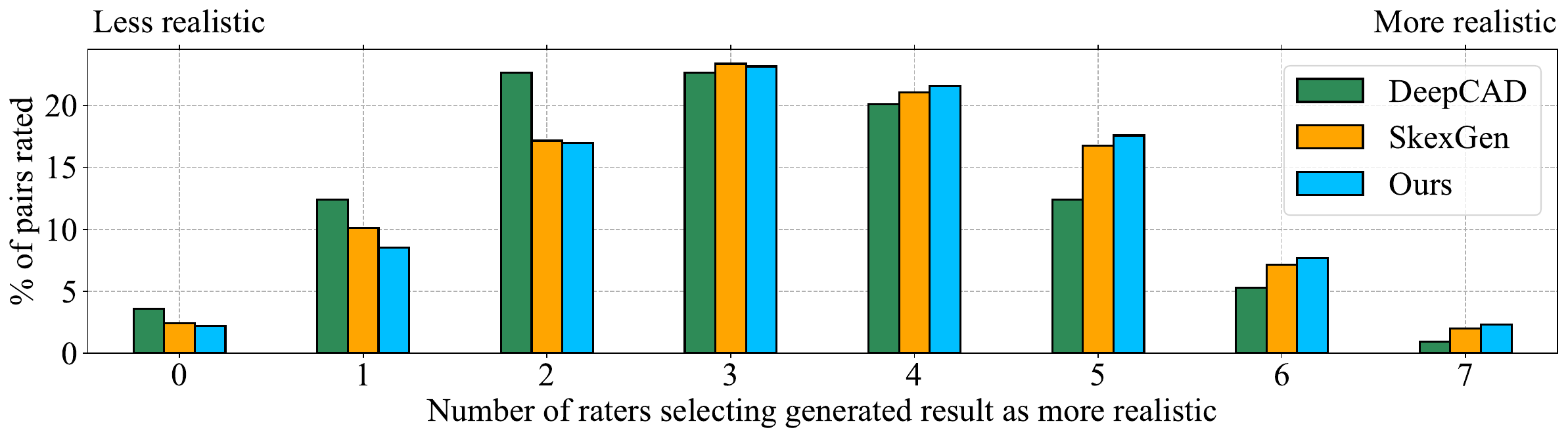}
    \caption{Distribution of votes by 7 human evaluators comparing the realism of complex samples produced by the three methods with the training set.}
    \label{fig:human_eval}
\end{figure}

\begin{figure}
    \centering
    \includegraphics[width=0.99\columnwidth]{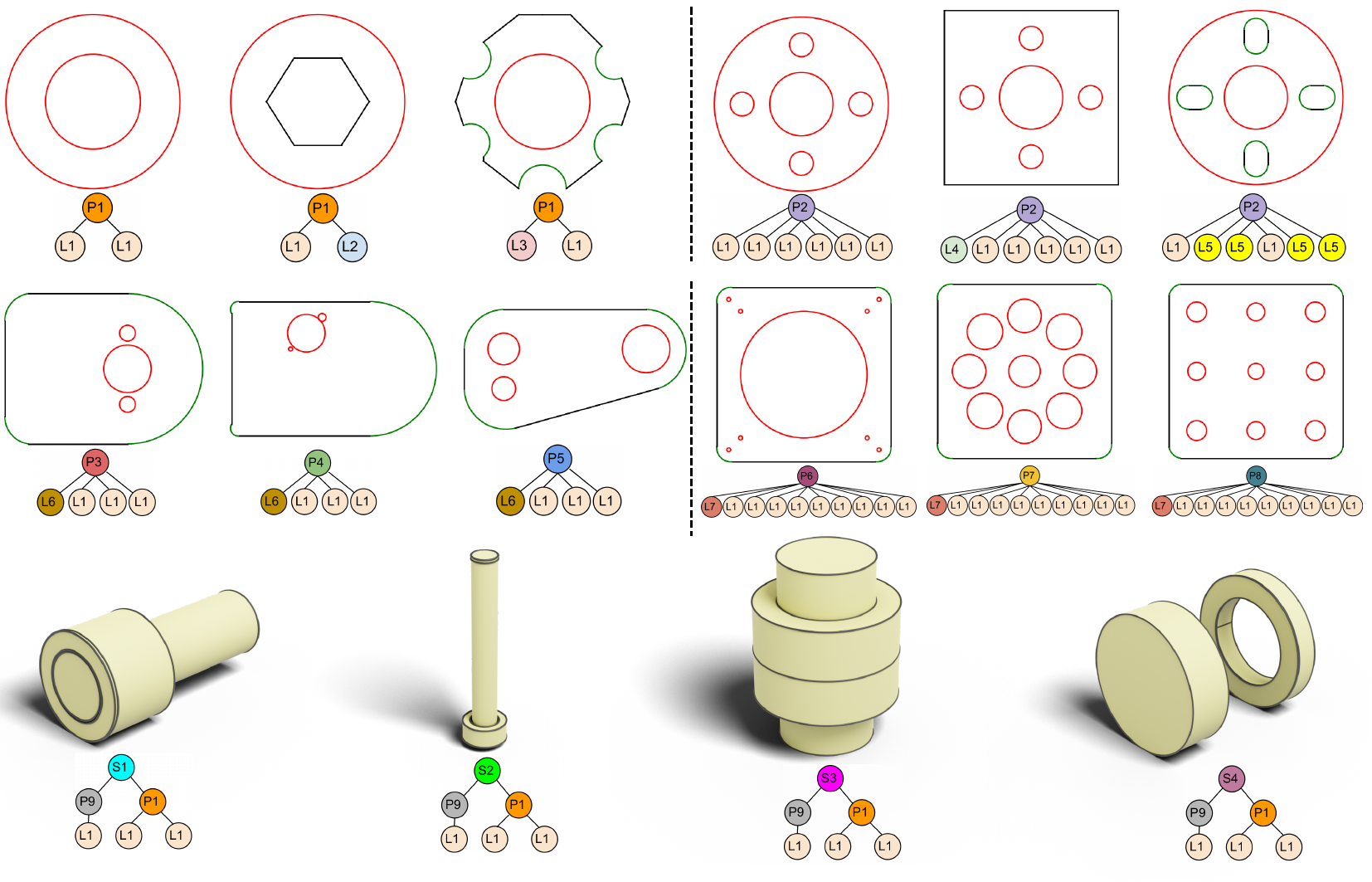}
    \caption{Generated results from hierarchical code tree editing. Code is edited in the (L)oop level of the tree in the first row, the (P)rofile level of the tree in the second row, and the (S)olid level of the tree in the third row. The code tree corresponding to each result is inset below.}
    \vskip -0.2in
    \label{fig:code_edit}
\end{figure}

\myparagraph{Human Evaluation}
To evaluate the perceived quality of our generation results, we run a human evaluation following the methodology in \cite{jayaraman2022solidgen}.   As our hierarchical technique excels at generating complex models, we choose to perform the human evaluation on models with three or more extrusions.  For the DeepCAD and SkexGen benchmarks, where control over the complexity of the generated models is not possible, we randomly sample models that have three or more extrusions from a larger pool of unconditional generation results. 
For each model created by a generative method, we randomly select another ground-truth model from DeepCAD and display renderings of the two side by side.  The image pairs were presented to crowd workers from the Amazon Mechanical Turk workforce \cite{mishra_2019}, who were asked to evaluate which of the two is more ``realistic''. To assist the crowd workers with this task, we provide carefully chosen examples of complex CAD models and low quality generations. See \autoref{sec:appendix_human_eval} for details.

Each image pair was rated independently by 7 crowd workers and we record the number of times generated data was selected as more realistic than training data, giving us a ``realism" score from 0 to 7. \autoref{fig:human_eval} shows the distribution of the``realism" scores.  We see that for our method the distribution is symmetric, indicating the crowd workers are unable to distinguish the generated models from the training.  DeepCAD and SkexGen distributions are skewed towards "less realistic", indicating crowd workers were able to identify models generated by them as simplistic or malformed.  We consider a generated model as more ``realistic" than the training data if 4 or more of the 7 raters selected it. For our method, 49.2\% of the generated models were more ``realistic" than complex examples from the training data compared to 46.9\% for SkexGen and 38.7\% for DeepCAD.

\begin{figure}[]
    \begin{center}
      \includegraphics[width=0.99\columnwidth]{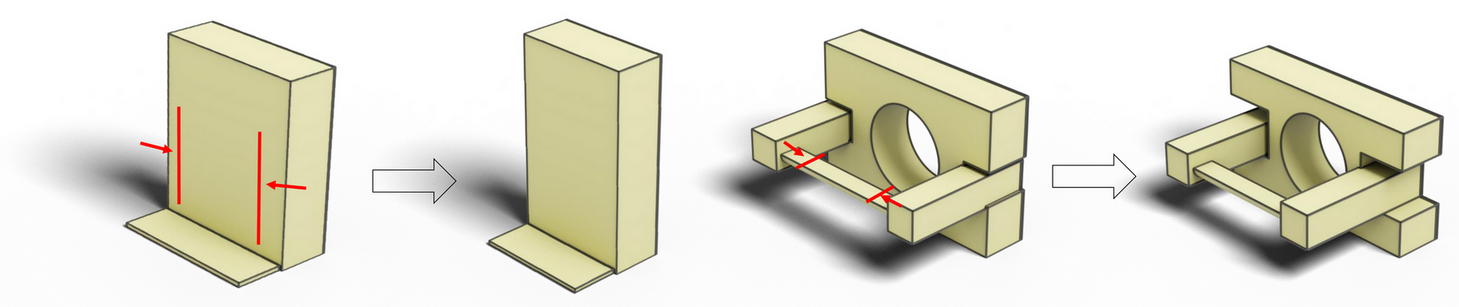}
        \caption{CAD parameter edits with fixed code trees. Red arrows indicate the individual parts edited by the user. Other parts automatically got modified.}
        \label{fig:cad_edit}
        \vskip -0.2in
    \end{center}
\end{figure}

\begin{figure}
    \begin{center}
      \includegraphics[width=0.99\columnwidth]{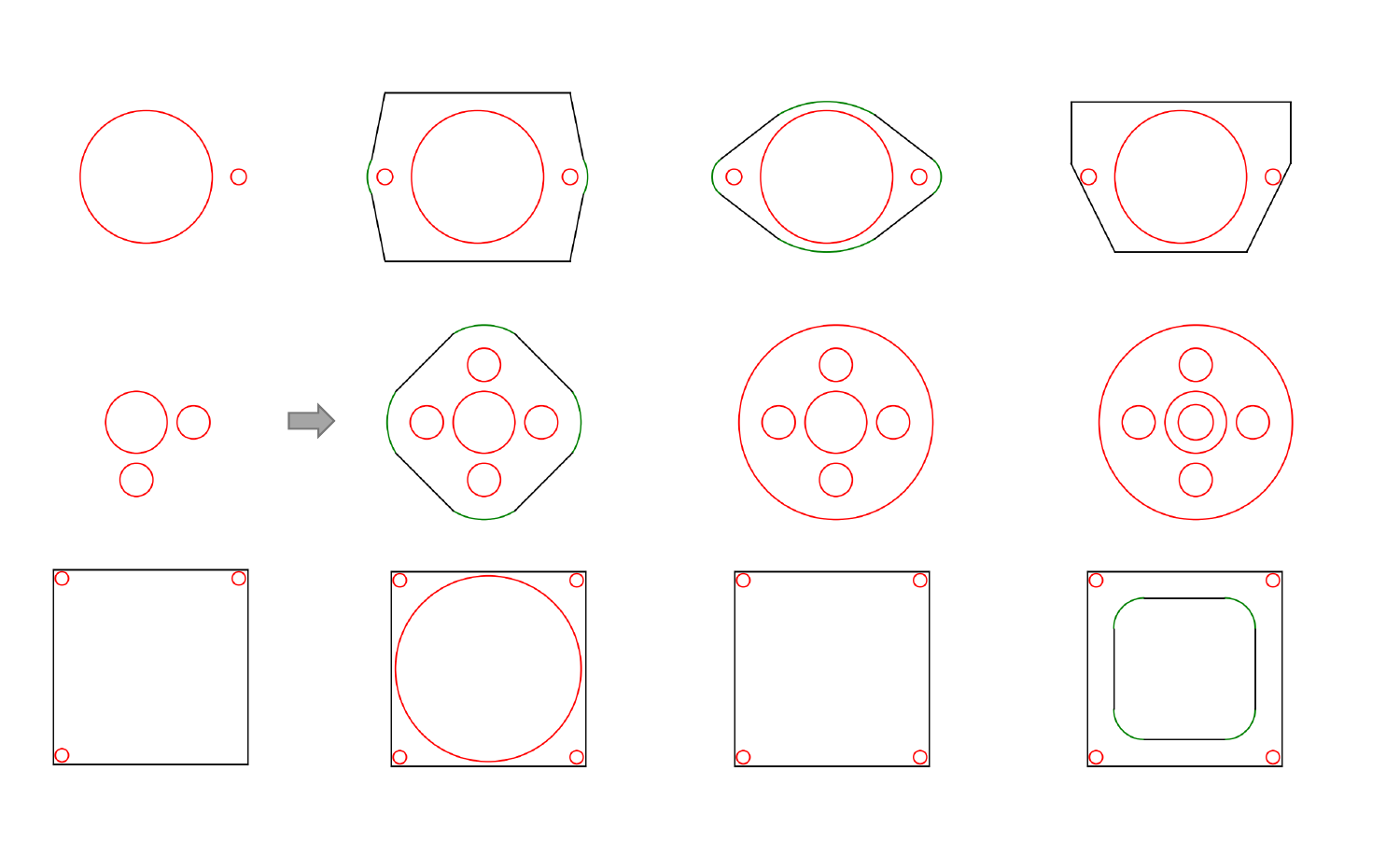}
        \caption{Autocompleted sketches (column 2 $\sim$ 4) from partial loops (column 1).}
        \label{fig:sketch_acfig}
        \vskip -0.2in
    \end{center}
\end{figure}

\subsection{Controllable Generation}
\label{sec:editing}

We demonstrate controllable generation in two ``editing'', and one ``auto-completion'' application scenarios.

\myparagraph{Code Tree Editing}
Given a code tree, a user can edit the code nodes at three different levels, achieving local and global modifications across the CAD hierarchy. This hierarchical control over the generation is unavailable in previous methods~\cite{wu2021deepcad,xu2022skexgen}. \autoref{fig:code_edit} illustrates the diverse and well-controlled generated results from editing of the code tree. We see that loop codes control the shape geometry, profile codes control the loop dimension and positioning in 2D, and finally solid code controls the height of extruded sketches and their 3D combination.

\begin{figure}[t]
    \begin{center}
    \begin{subfigure}[t]{\columnwidth}
    \centering
    \includegraphics[width=\textwidth]{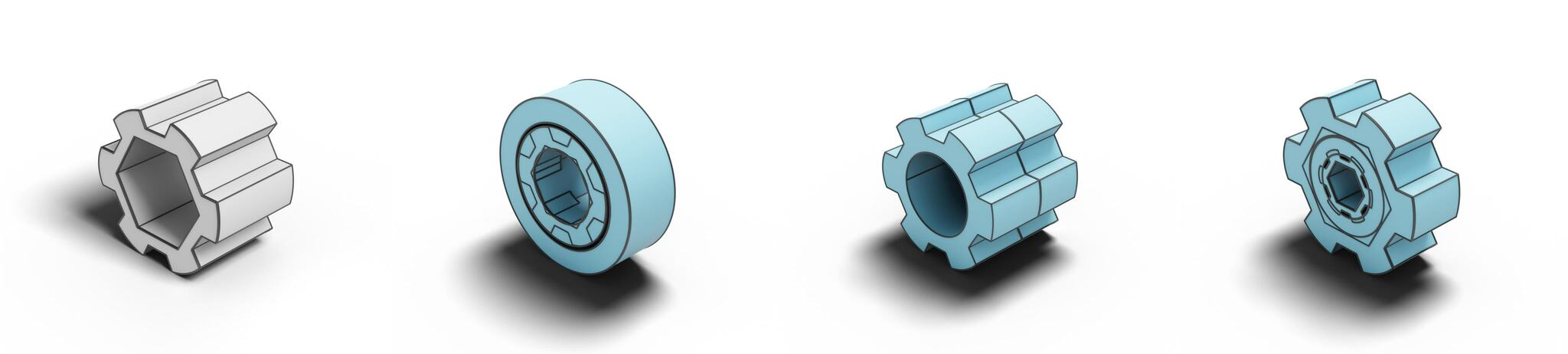}
    \end{subfigure}\\
    \vskip -0.02in
    \begin{subfigure}[t]{\columnwidth}
    \centering
    \includegraphics[width=\textwidth]{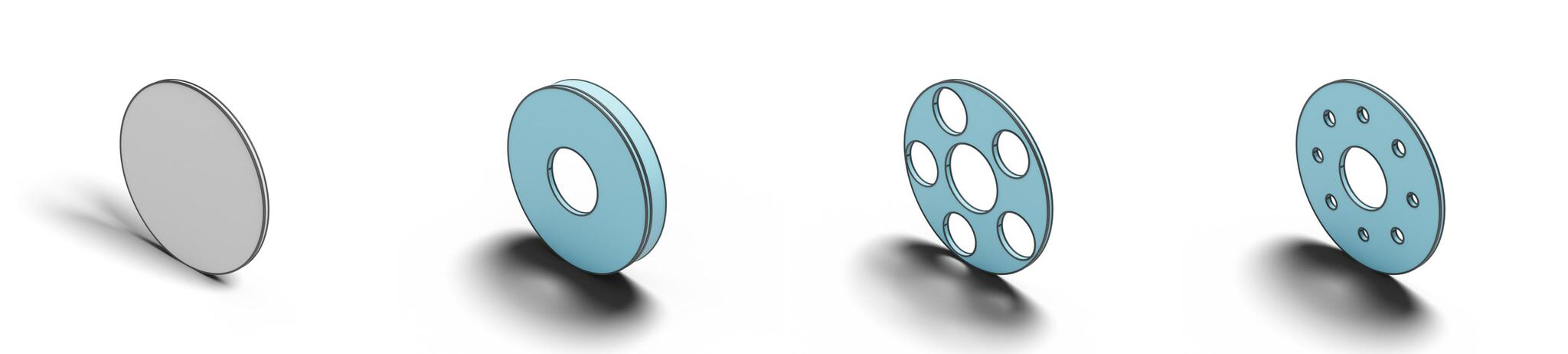}
    \end{subfigure}
    \vskip -0.02in
    \begin{subfigure}[t]{\columnwidth}
    \centering
    \includegraphics[width=\textwidth]{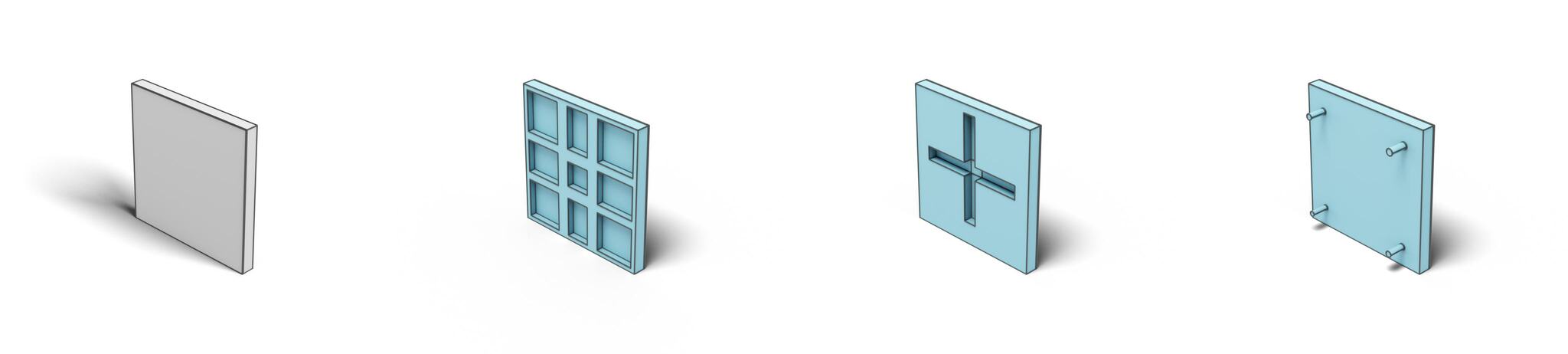}
    \end{subfigure}\\
    \vskip -0.02in
    \begin{subfigure}[t]{\columnwidth}
    \centering
    \includegraphics[width=\textwidth]{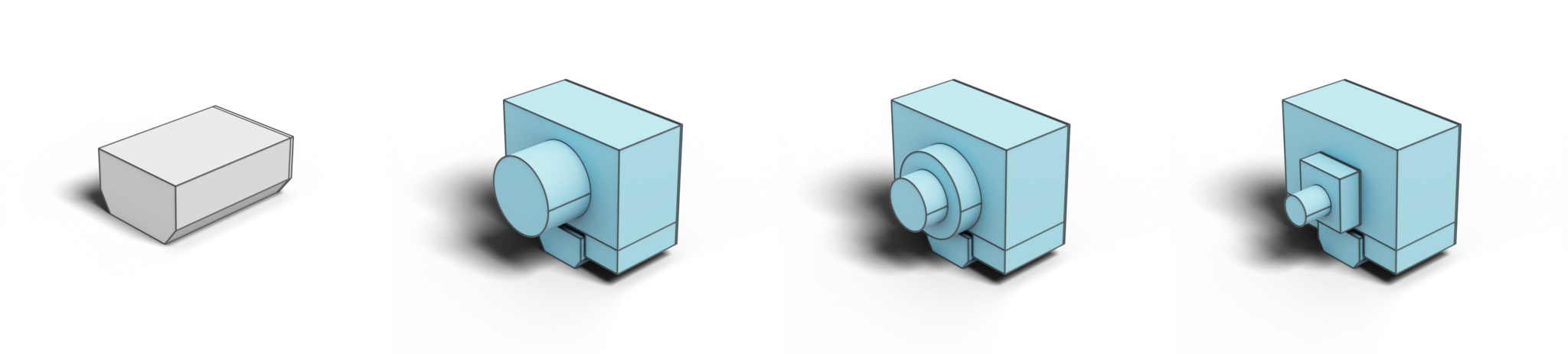}
    \end{subfigure} \\
    \vskip -0.02in
    \begin{subfigure}[t]{\columnwidth}
    \centering
    \includegraphics[width=0.9\textwidth]{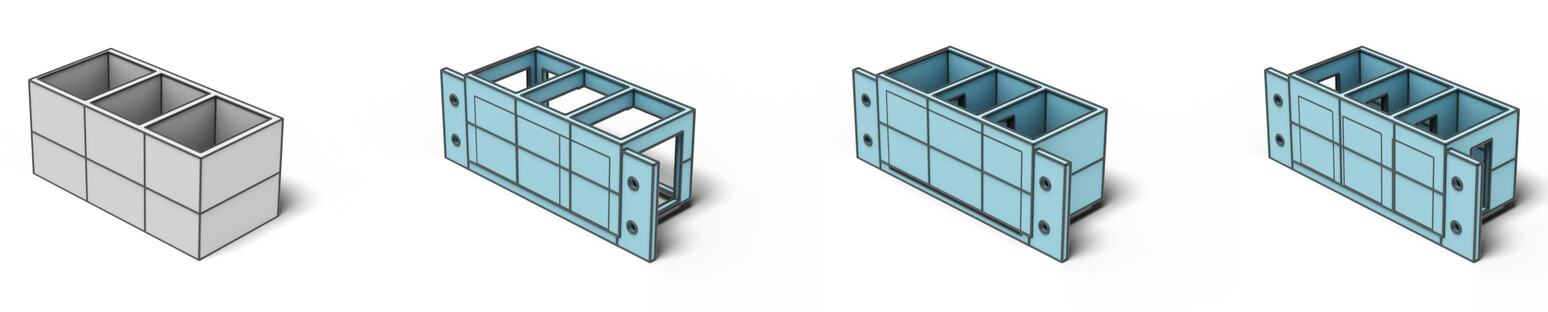}
    \end{subfigure}
    \caption{Autocompleted CAD models (blue) from partial extruded profiles (gray).}
    \vskip -0.2in
    \label{fig:cad_autocomplete}
    \end{center}
\end{figure}

\myparagraph{Design-Preserving Editing}
With the code tree fixed, a user can preserve the current design while making local edits to the model parameters to iteratively refine it. Treating user edited parameters as partial input and reuse the previous neural codes, the model generator outputs a new CAD model following both the current design and the user edited values. \autoref{fig:cad_edit} demonstrates that after a user edit to the horizontal length of a local part, the bottom part in the left and the two supporting arms in the right adjusted their size accordingly to accommodate the user edit. This automatic process is the result of keeping the code tree that encodes the part connectivity and dimension relations. 

\begin{figure*}
    \begin{center}
        \includegraphics[width=\textwidth]{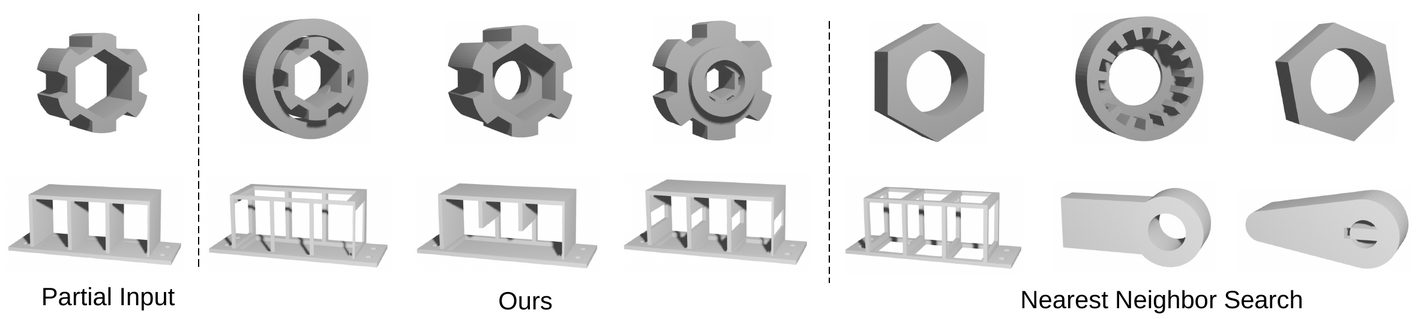}
        \caption{Qualitative comparison between our method (center) and the nearest neighbor search baseline (right). Given the same partial user input (left), our method autocompletes the CAD model with better diversity and fidelity.}
        \label{fig:knn_compare}
    \end{center}
    \vskip -0.1in
\end{figure*}

\myparagraph{Autocompletion from User Input}
We consider partial user input in the format of one or multiple extruded profiles or loops. Our code tree generator can predict a set of likely codes from partial input and use the generated code together with the partial input to autocomplete the full CAD model. \autoref{fig:sketch_acfig} shows the sketch autocomplete results when a user provide partial loops and model completes the full sketch. Likewise, \autoref{fig:cad_autocomplete} shows the CAD autocomplete results from partial extruded profiles. Each row contains multiple generated results from different generated codes. Here we use top-1 sampling in the model generator to limit the generation diversity to code only. See \autoref{sec:appendix_cad_results} and \autoref{sec:appendix_sketch_results} for additional results. 

For comparison, we also implemented a nearest-neighbor search baseline. Partial CAD solids built from intermediate steps of the sketch-and-extrude formed a ground-truth incomplete CAD database. 
User input is the query and we compute its Chamfer distance to all shapes in the database. The k-nearest shapes are considered to be geometrically similar and we retrieve their corresponding ground-truth complete CAD models as the completed result. 
\autoref{fig:knn_compare} compares our generated results with the top-3 nearest neighbor results. Nearest-neighbor results have less diversity and fail to closely match the user input. In contrast, our results correctly auto-complete the user input with high diversity.

\subsection{Instance-Agnostic Design Pattern} 
\begin{figure}
    \begin{center} \includegraphics[width=0.99\columnwidth]{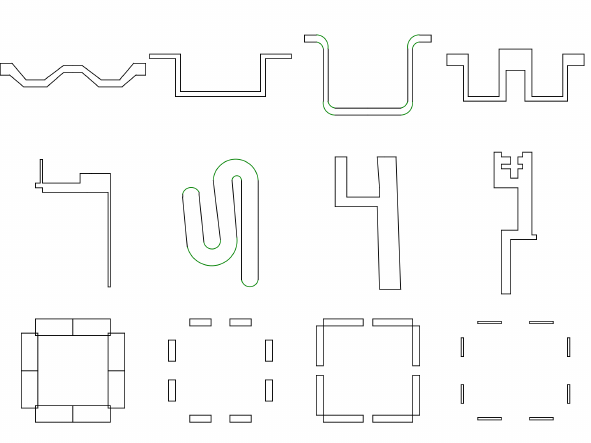}
        \caption{Loops (row 1,2) and profiles (row 3) encoded to the same code. Profiles shown as bounding boxes.}
        \label{fig:codebook_cluster}
        \vskip -0.2in
    \end{center}
\end{figure}

To better understand the unsupervised features learned by the codebook, \autoref{fig:codebook_cluster} shows the data and code mapping after encoding. We see that data assigned to the same code share similar instance-agnostic design patterns, such as the oscillating pattern in the first row, while effectively ignoring data-specific details like the exact number of curves or its type. More visualization is in \autoref{sec:appendix_code_code_mapping}.

\section{Limitations}
\label{sec:limitation}
A primary failure mode of our current system is the lack of validity in the generated CAD models with self-intersecting edges or solids. Our loss functions do not explicitly penalize invalid geometries; future work is the addition of a loss function that explicitly penalizes the CAD model invalidity with domain knowledge. Another direction is to learn to recover from such failures, which currently poses a challenge due to the lack of an ``invalid CAD model dataset''. Lastly, another limitation of our approach is the use of the sketch-and-extrude CAD format that excludes other popular modeling operations such as revolve, mirror, and sweep.

\section{Conclusion}
\label{sec:conclude}
We introduce a novel generative model for controllable CAD generation. A key to our approach is a three-level neural coding that captures design patterns and intent at different levels of the modeling hierarchy. This paper makes another step towards intelligent generative design with users in the loop. Extensive evaluations demonstrate major boosts in generation quality and promising applications of our hierarchical neural coding such as intent-aware editing or auto-completion. 
%

\section*{Acknowledgements}
This research is partially supported by NSERC Discovery Grants with Accelerator Supplements and DND/NSERC Discovery Grant Supplement, NSERC Alliance Grants, and John R. Evans Leaders Fund (JELF).

\bibliography{main}
\bibliographystyle{icml2023}

\newpage
\appendix
\onecolumn

\section{Human Evaluation Details}
\label{sec:appendix_human_eval}
\begin{figure}
    \centering
    \newcommand\imgwidth{0.12\columnwidth}
    
    \begin{tabular}{ c c c c c c }

    \includegraphics[width=\imgwidth, trim=80 0 80 0, clip]{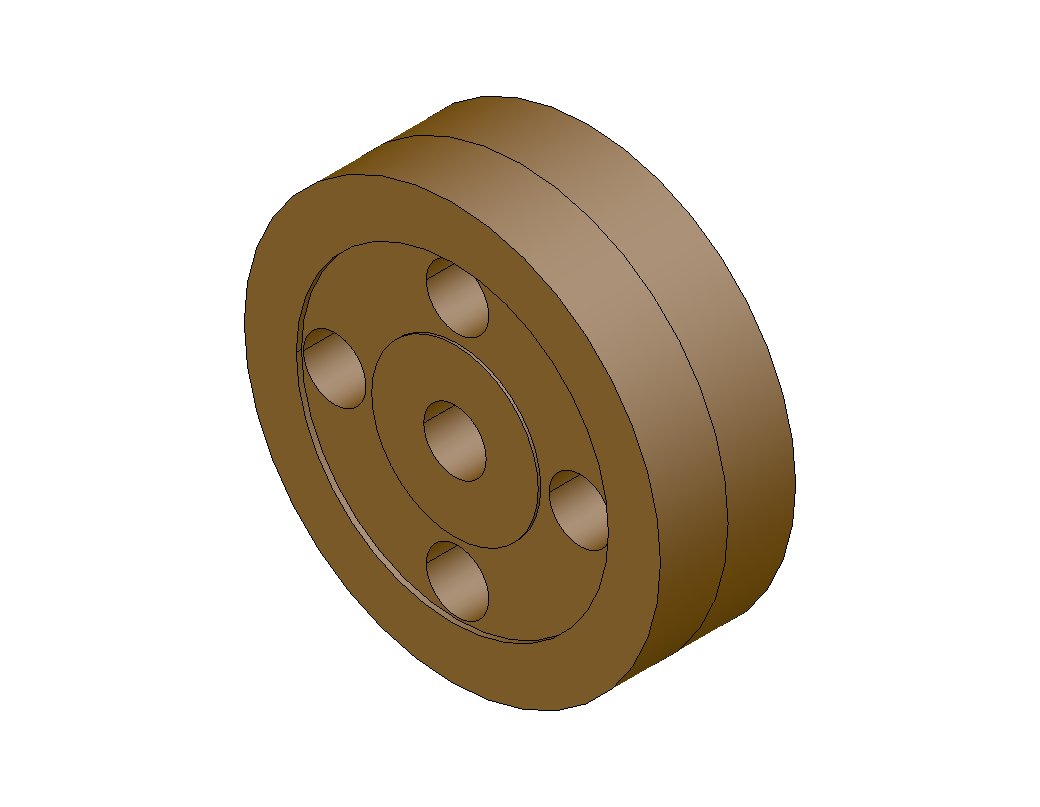} &
    \includegraphics[width=\imgwidth, trim=30 0 30 0, clip]{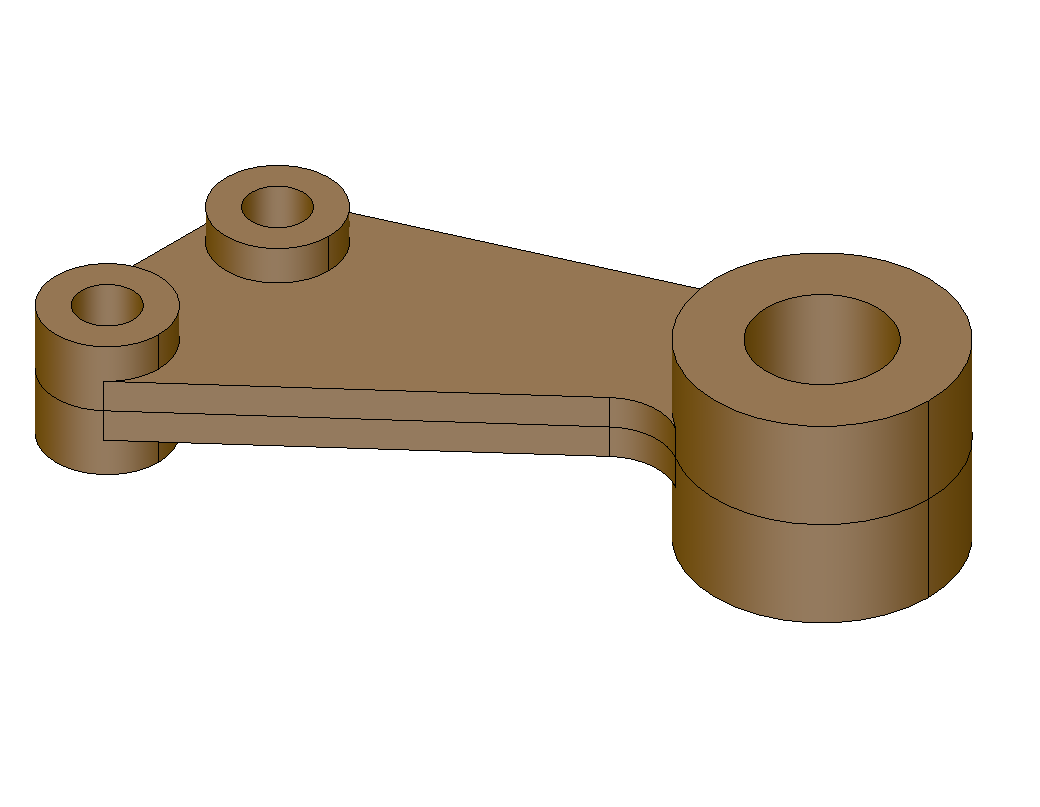} &
    \includegraphics[width=\imgwidth, trim=30 0 30 0, clip]{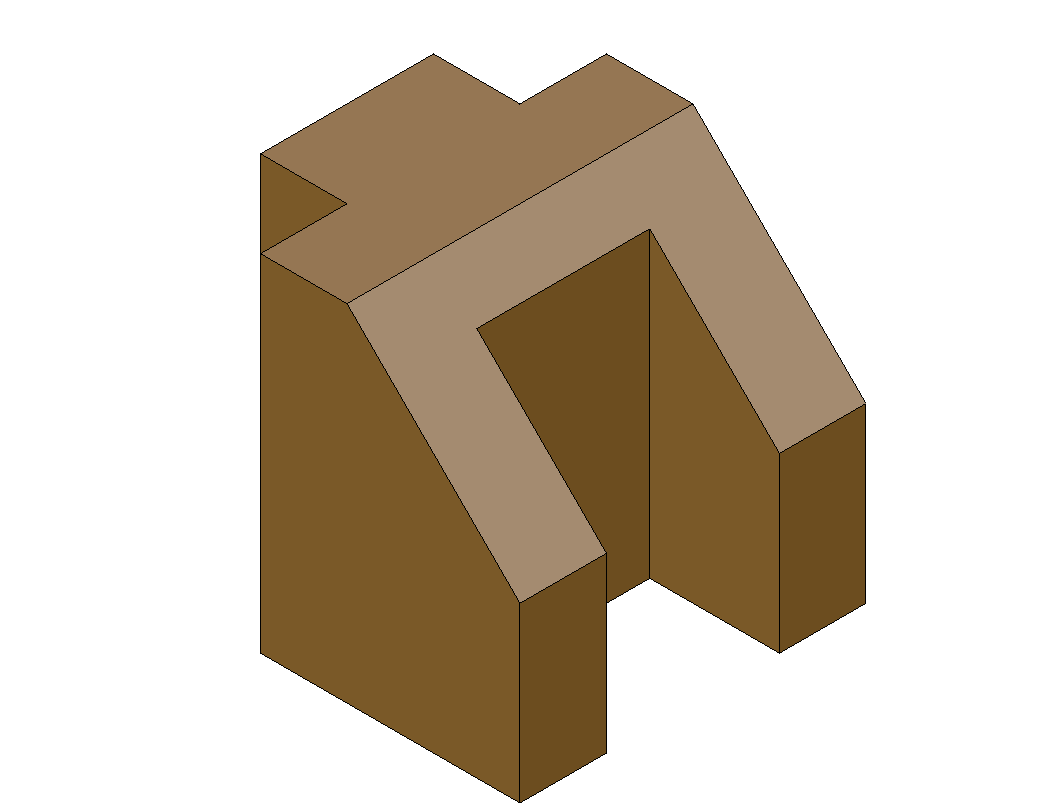} &
    \includegraphics[width=\imgwidth, trim=80 0 80 0, clip]{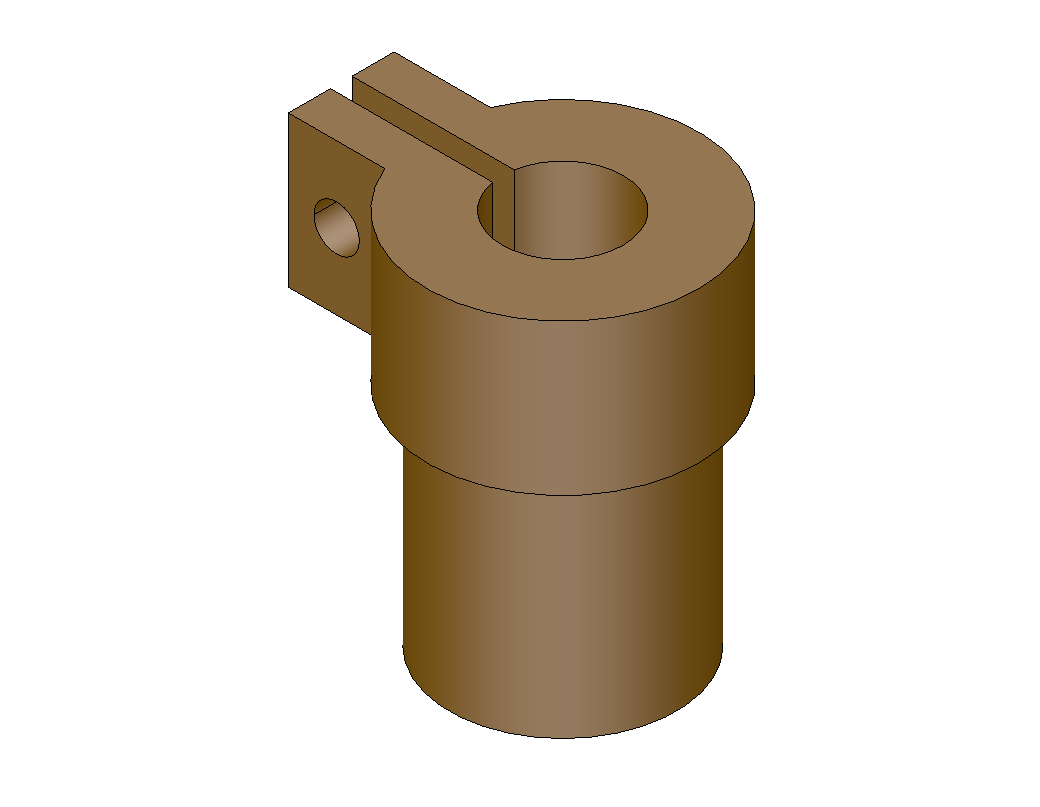} &
    \includegraphics[width=\imgwidth, trim=10 0 10 0, clip]{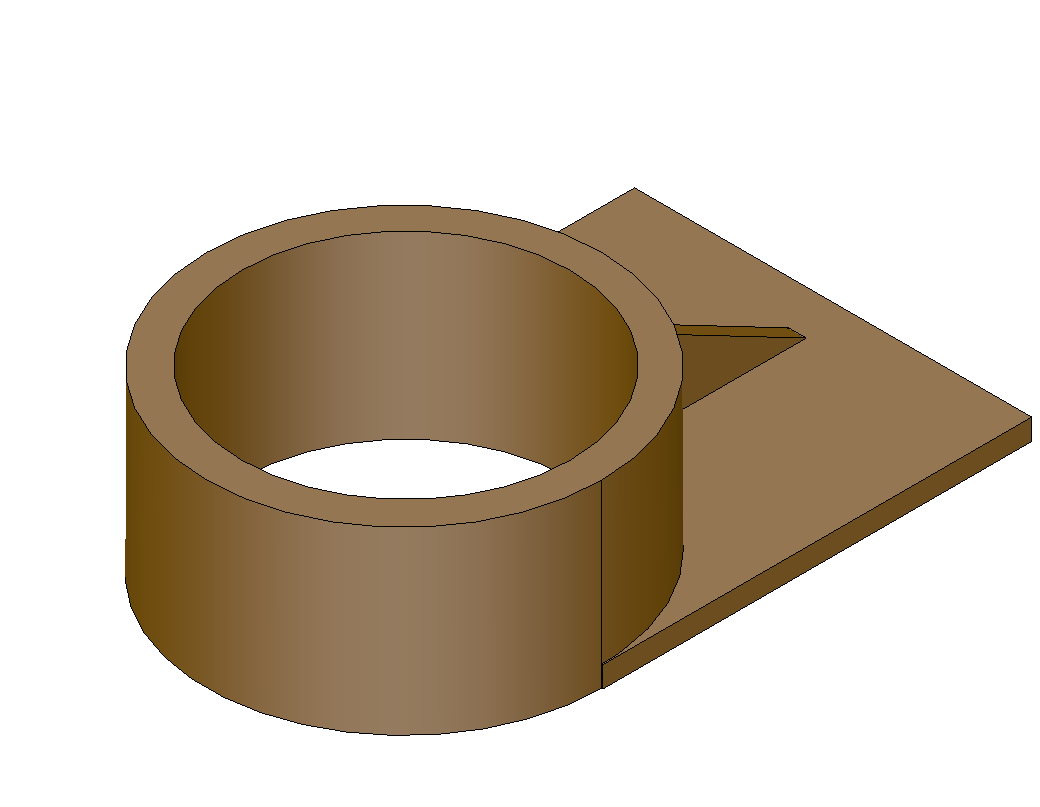} &
    \includegraphics[width=\imgwidth, trim=80 0 80 0, clip]{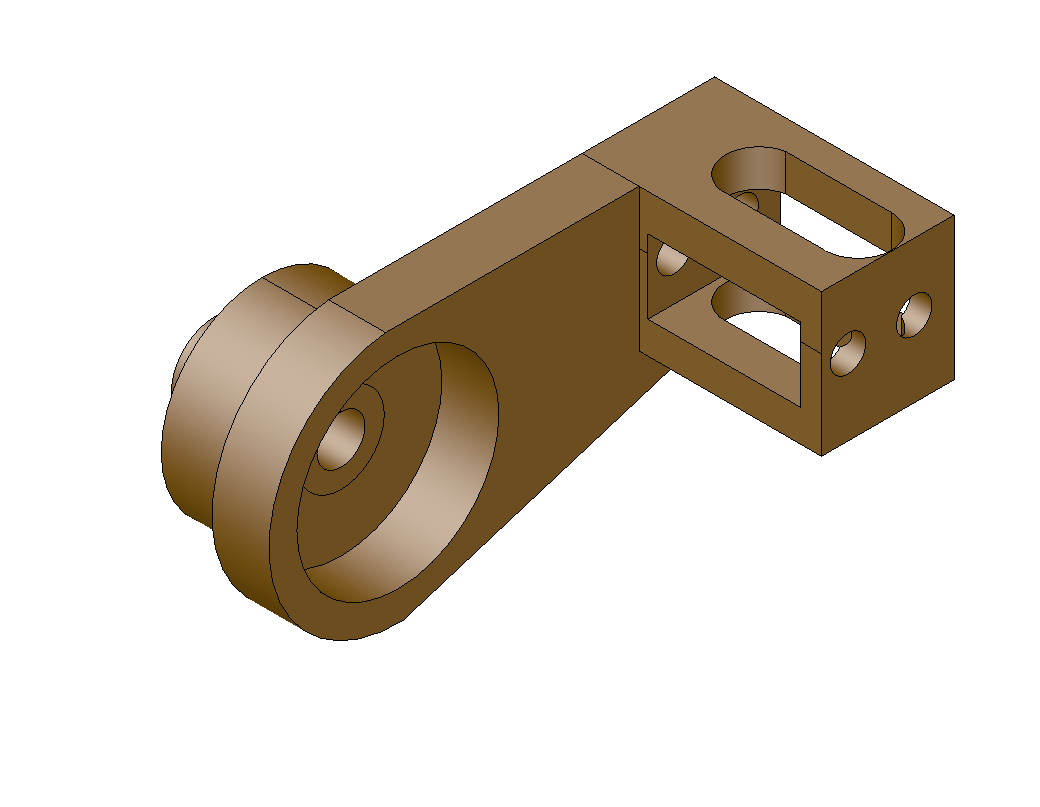} \\
    \multicolumn{6}{c}{Realistic models}
    \end{tabular}
    \begin{tabular}{ c c c c c }
    \includegraphics[width=\imgwidth, trim=30 0 30 0, clip]{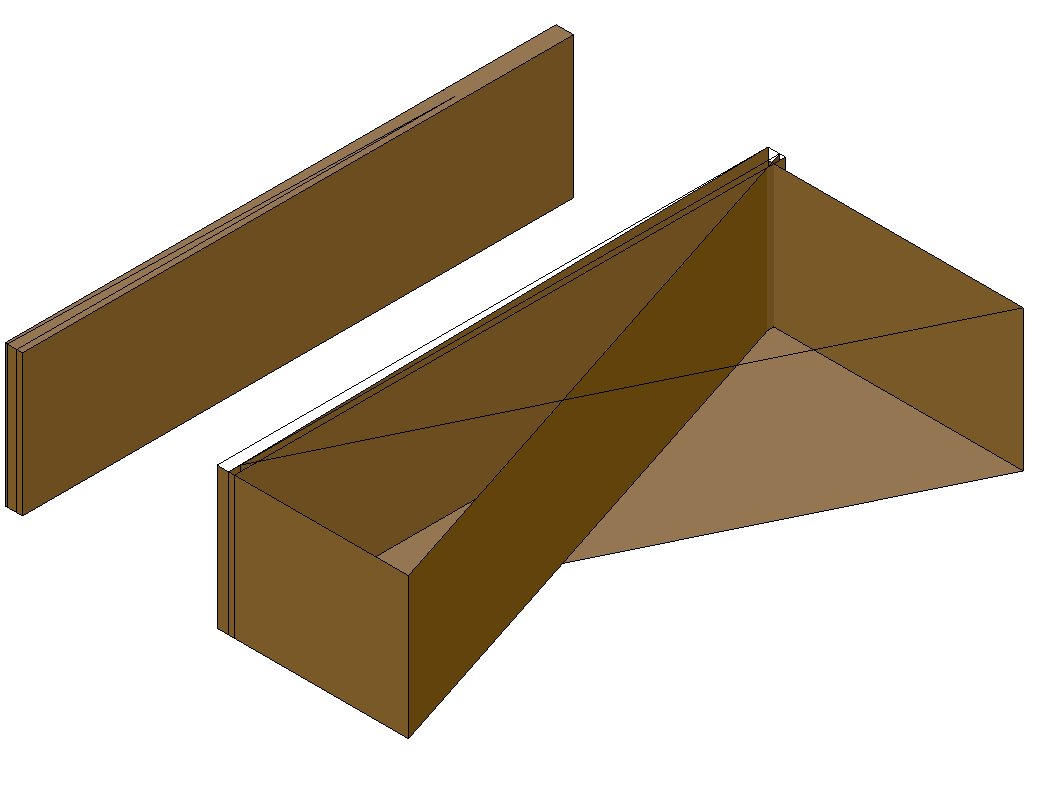} &
    \includegraphics[width=\imgwidth, trim=80 0 80 0, clip]{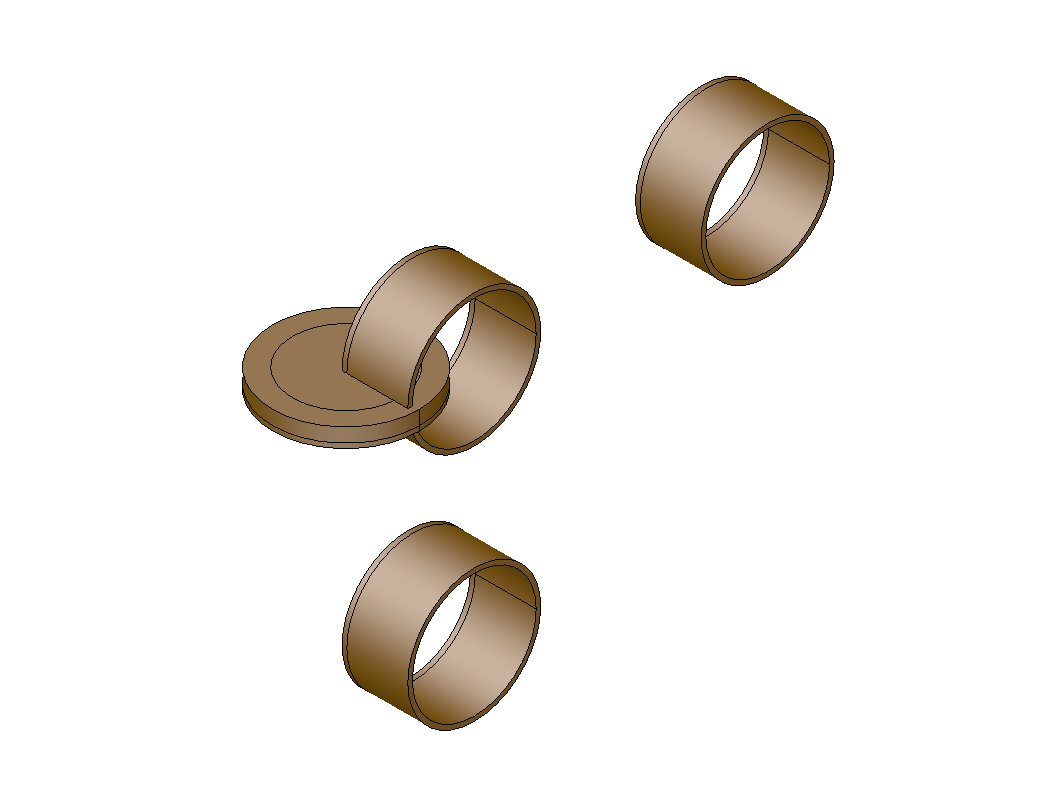} &
    \includegraphics[width=\imgwidth, trim=80 0 80 0, clip]{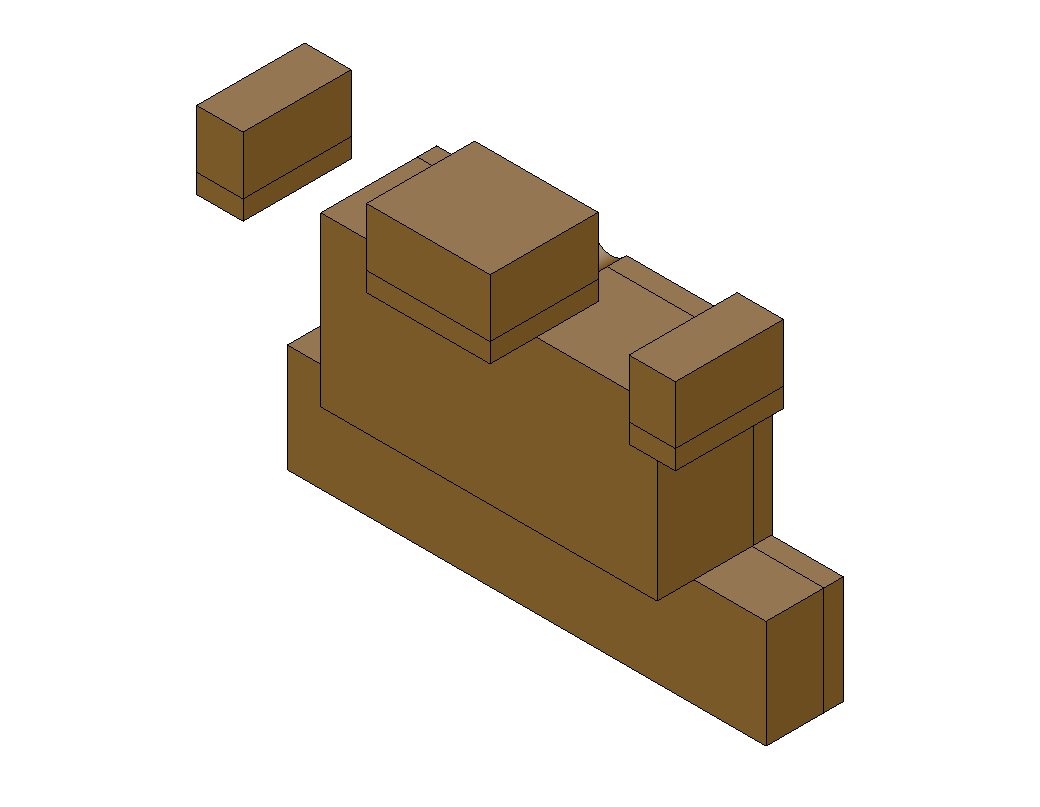} &
    \includegraphics[width=\imgwidth, trim=80 0 80 0, clip]{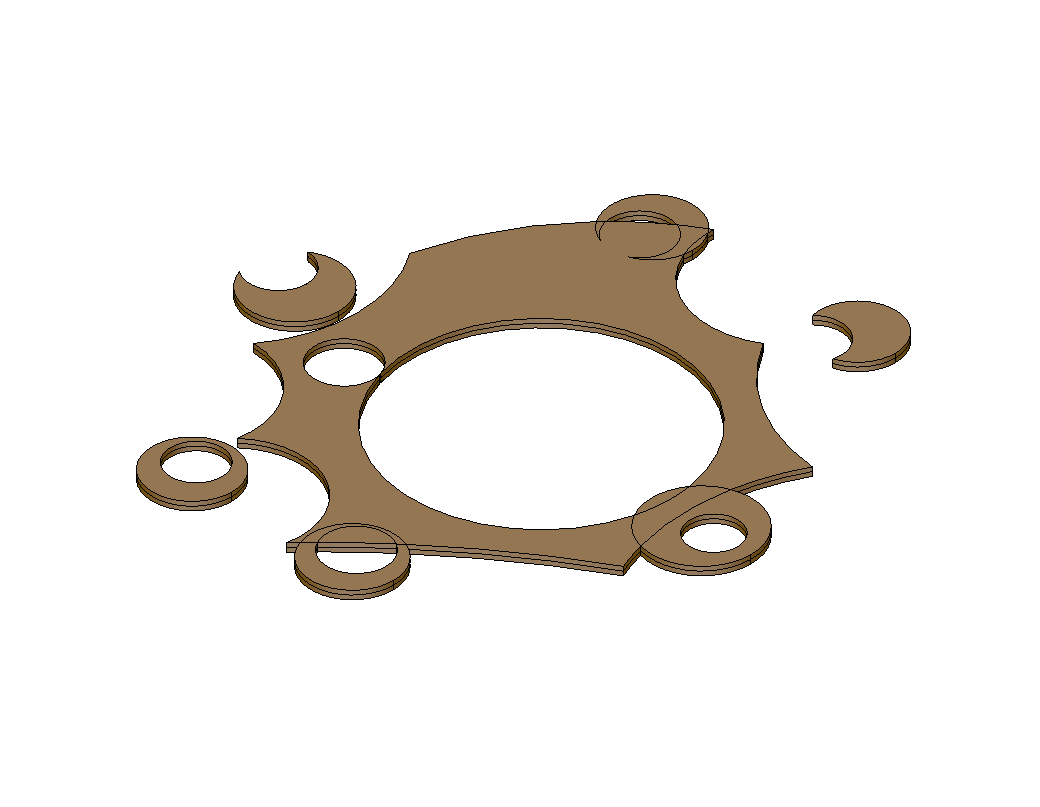} &
    \includegraphics[width=\imgwidth, trim=80 0 80 0, clip]{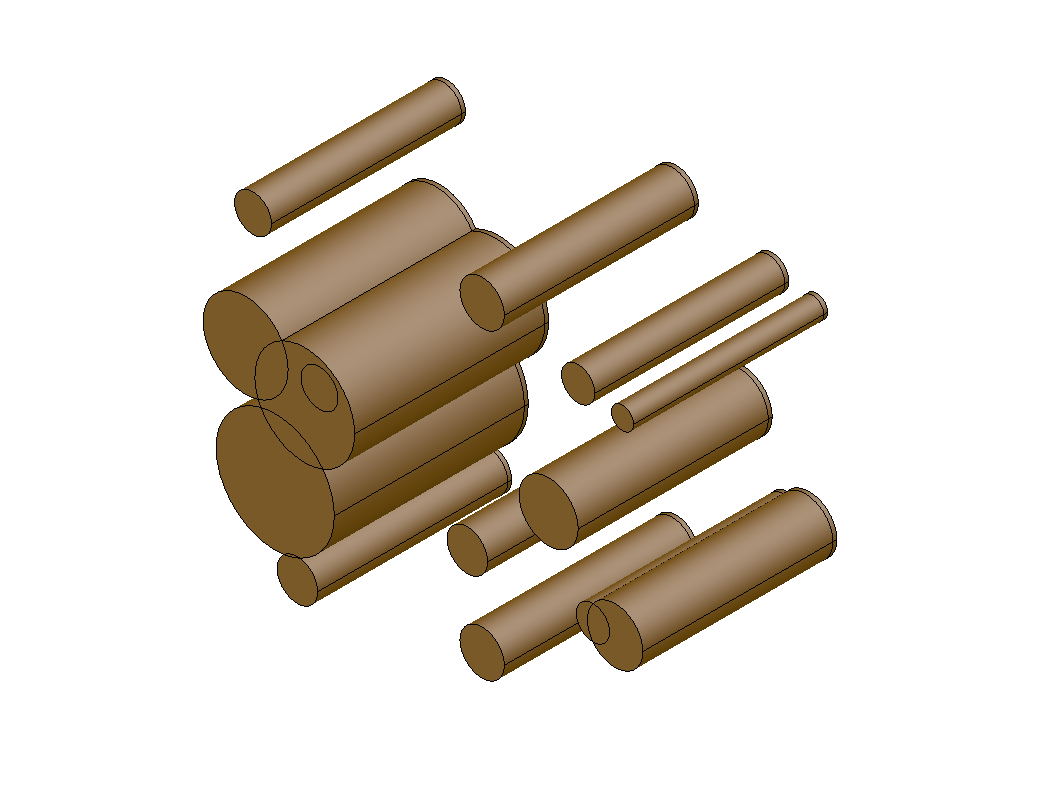}  \\
    \includegraphics[width=\imgwidth, trim=80 0 80 0, clip]{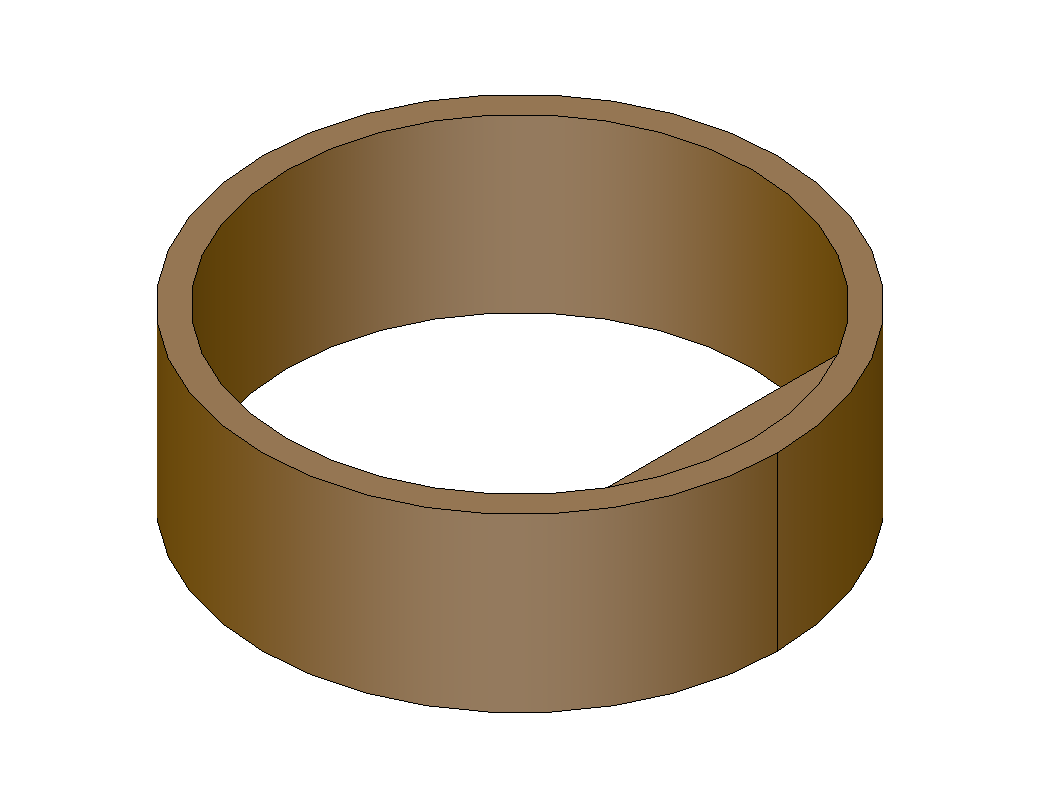} &
    \includegraphics[width=\imgwidth, trim=80 0 80 0, clip]{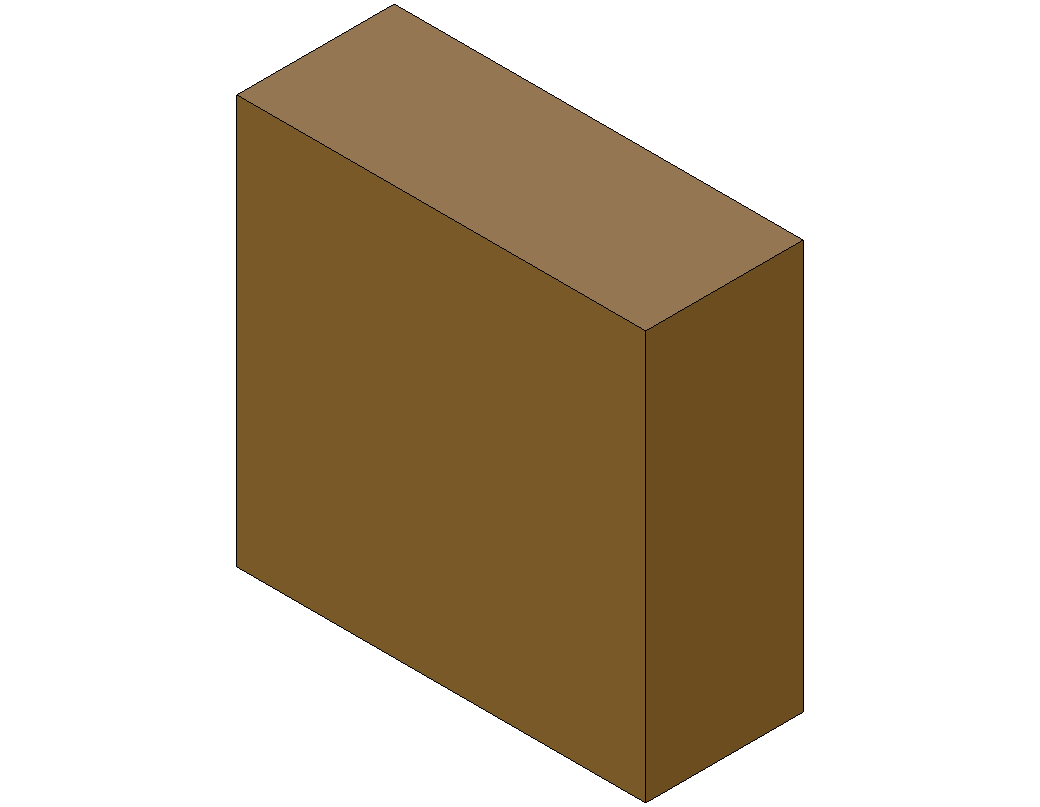} &
    \includegraphics[width=\imgwidth, trim=80 0 80 0, clip]{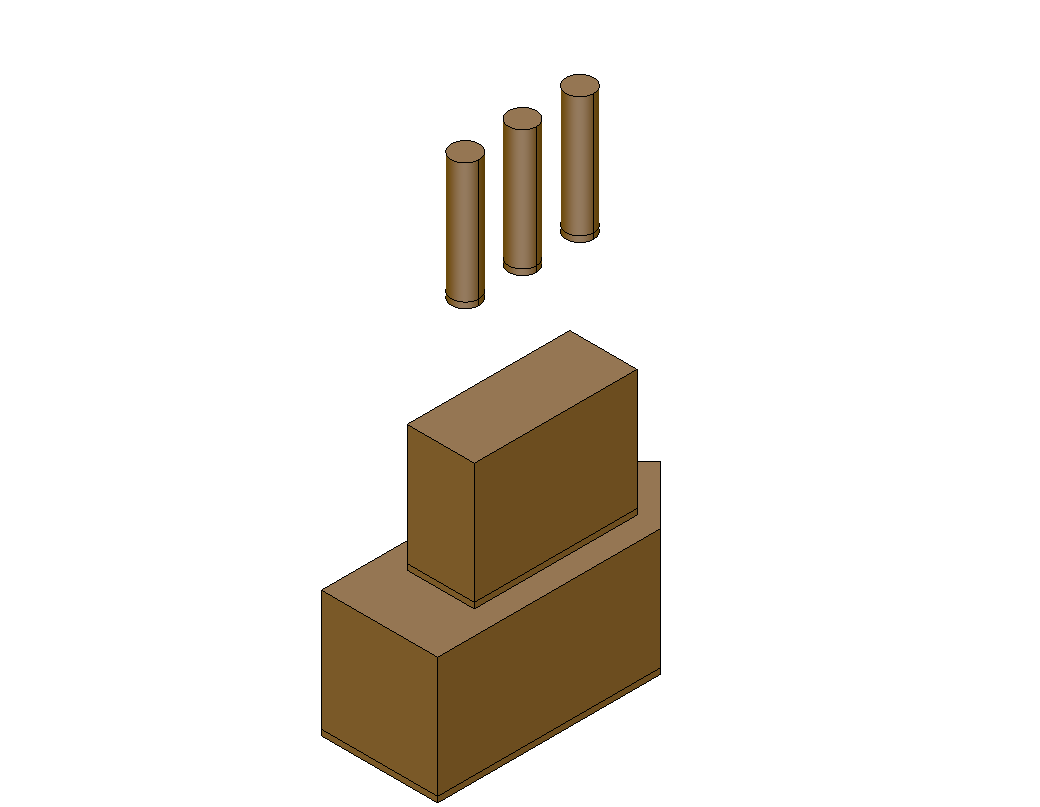} &
    \includegraphics[width=\imgwidth, trim=80 0 80 0, clip]{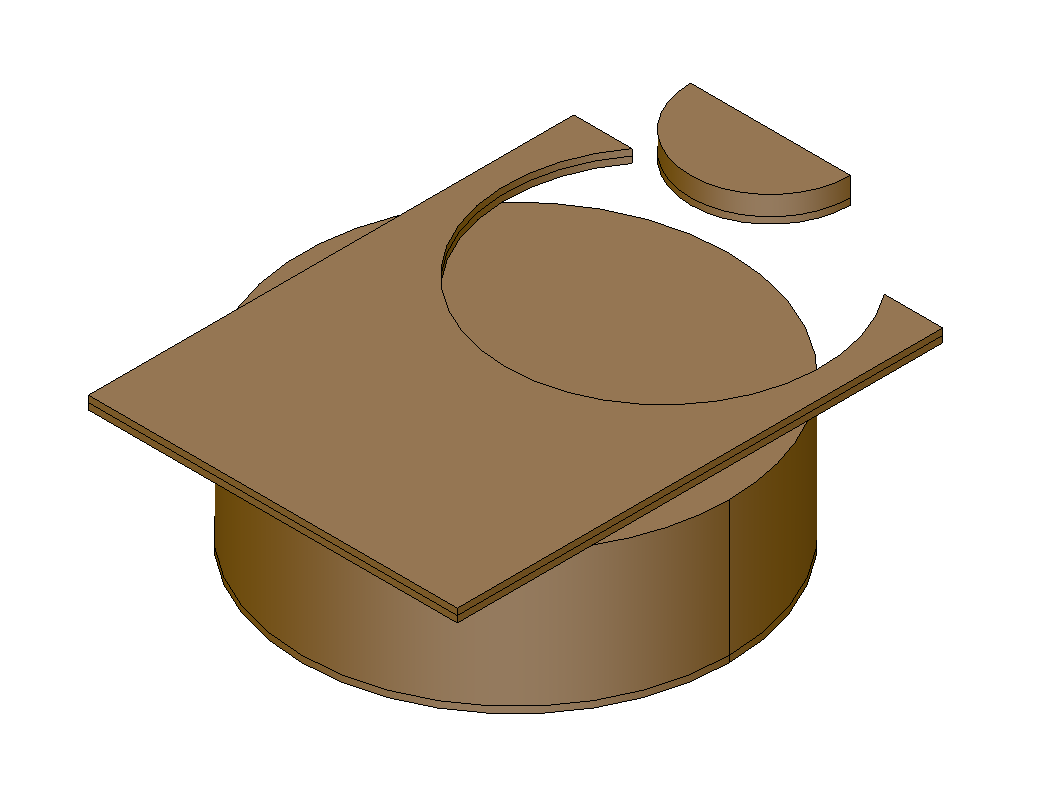} &
    \includegraphics[width=\imgwidth, trim=80 0 80 0, clip]{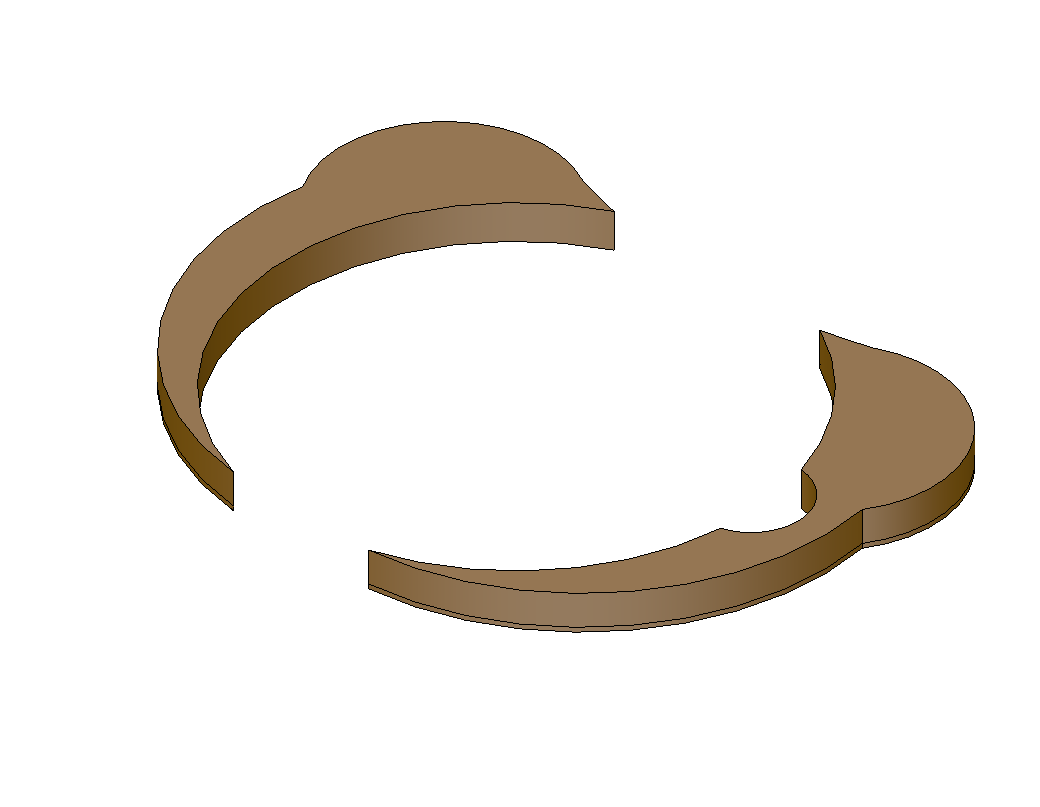} \\
    \multicolumn{5}{c}{Unrealistic models}
    \end{tabular}
    \caption{Images shown to crowd workers of realistic and unrealistic models.}
    \label{fig:realistic_and_unrealistic_solids}
\end{figure}

The perceived quality of our generation results was assessed using a ``two-alternative forced choice" task, in which crowd workers from the Amazon Mechanical Turk workforce were asked to choose whether a generated model was more ``realistic'' than a random model from training data.    To assist the crowd workers, we provide examples of six high quality models from the DeepCAD training data and ten examples of bad quality generations.  The example images are shown in \autoref{fig:realistic_and_unrealistic_solids}.  The high quality models are chosen to include desirable properties like symmetry, complexity and a coherent design, while the low quality examples include models which are very simple, not watertight or incoherent jumbles of extrusions.  The evaluation was conduced on 950 image pairs for each generative method, with 7 crowd workers rating each pair, resulting in a total of 19,950 human annotations.

\section{Additional CAD Results}
\label{sec:appendix_cad_results}

\begin{figure*}
    \centering
    \includegraphics[width=0.7\textwidth]{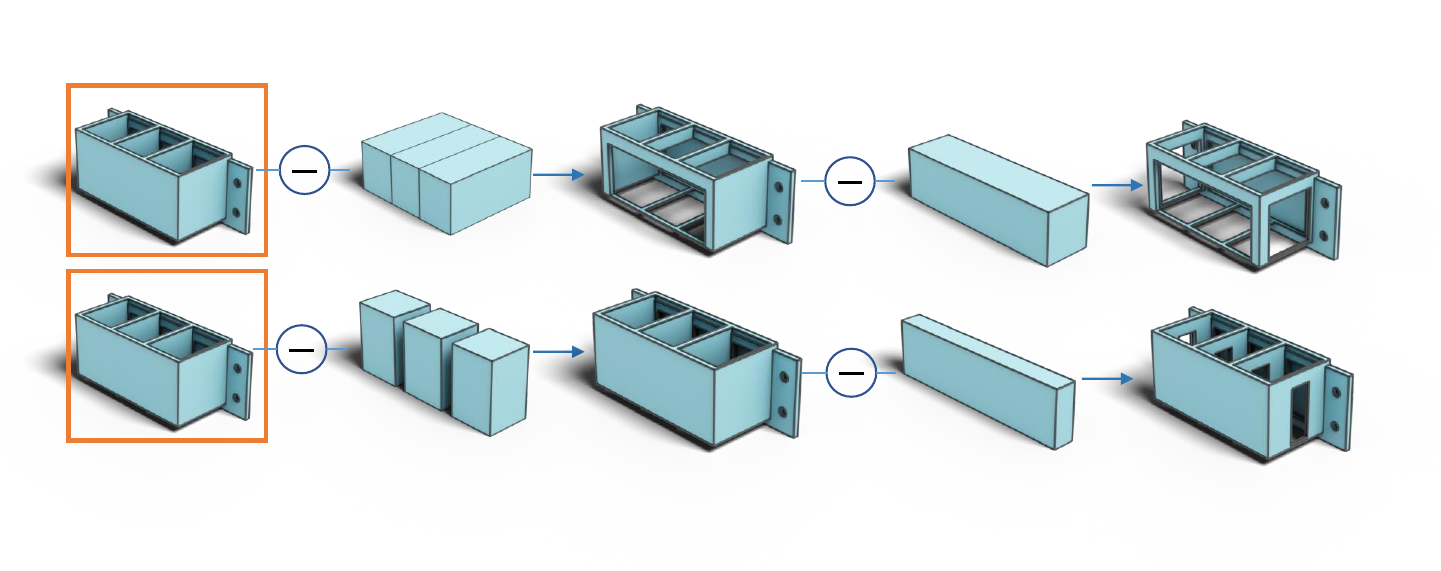}
    \includegraphics[width=0.7\textwidth]{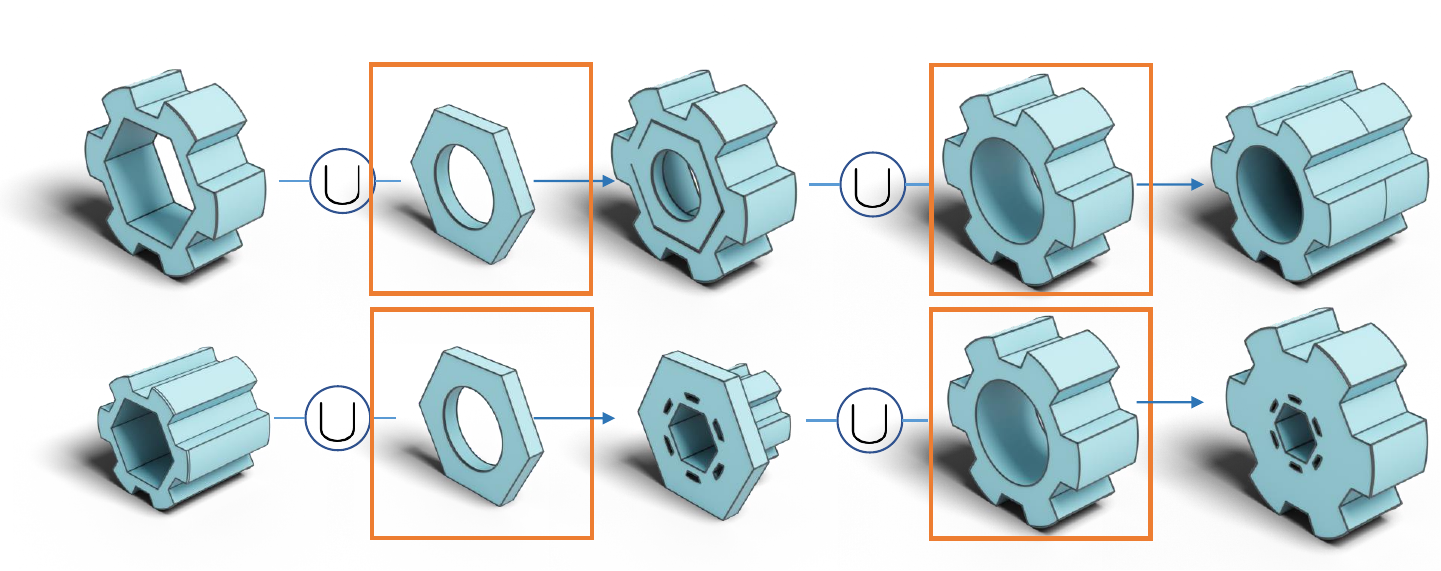}
    \caption{Autocompleted CAD models from partial user input. The order of sketch-and-extrude is from left to right. User provided extruded profiles are colored in orange. We show two different autocompletion results from the same input. In the second example, user input consists of multiple extruded profiles at different steps of the construction sequence.}
    \label{fig:cad_ac_detailed}
\end{figure*}

\begin{figure*}
    \centering
    \includegraphics[width=0.9\textwidth]{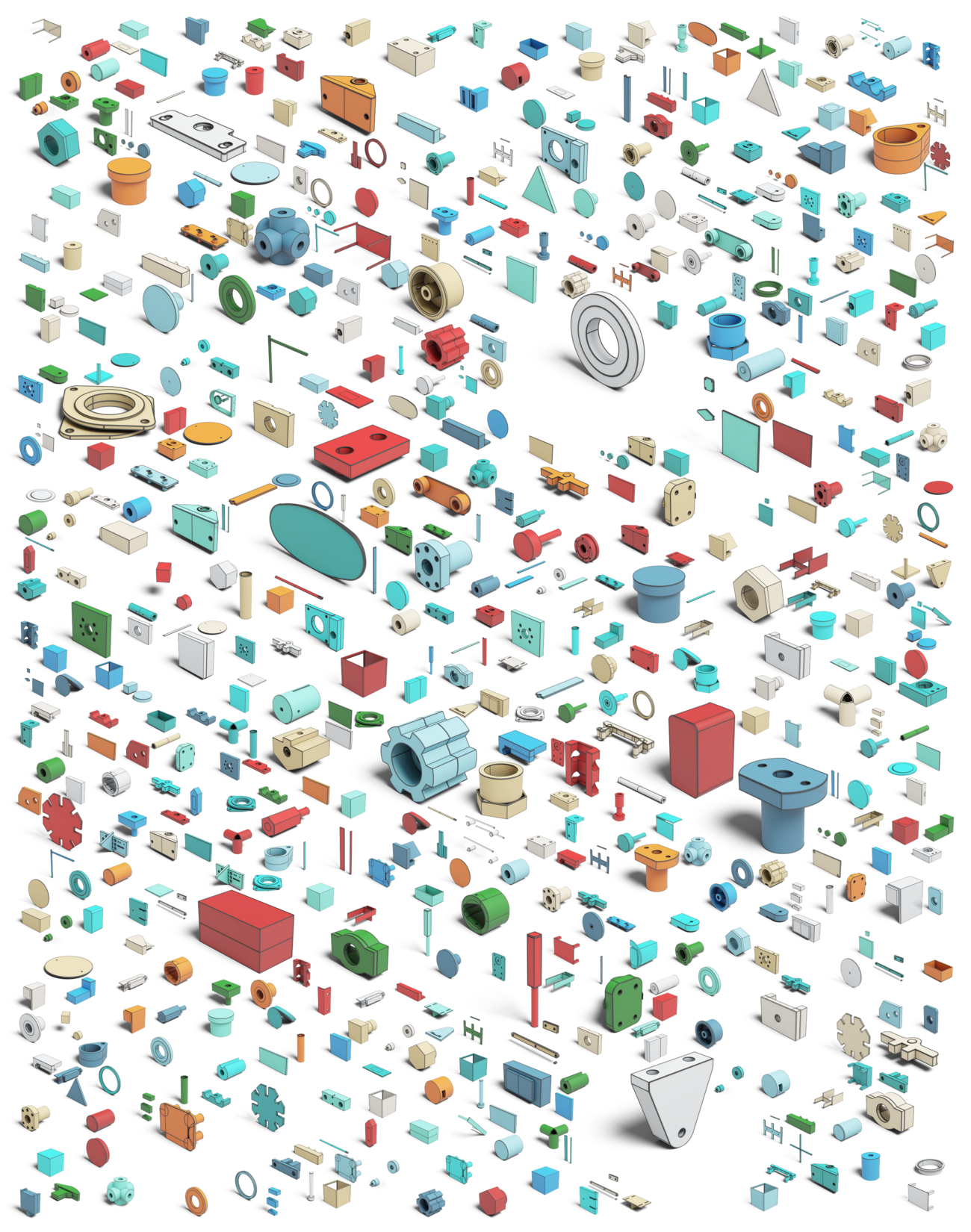}
    \caption{Randomly generated CAD models from our method.}
    \label{fig:mosaic_appendix}
\end{figure*}

\begin{figure*}
    \centering
    \includegraphics[width=0.95\textwidth]{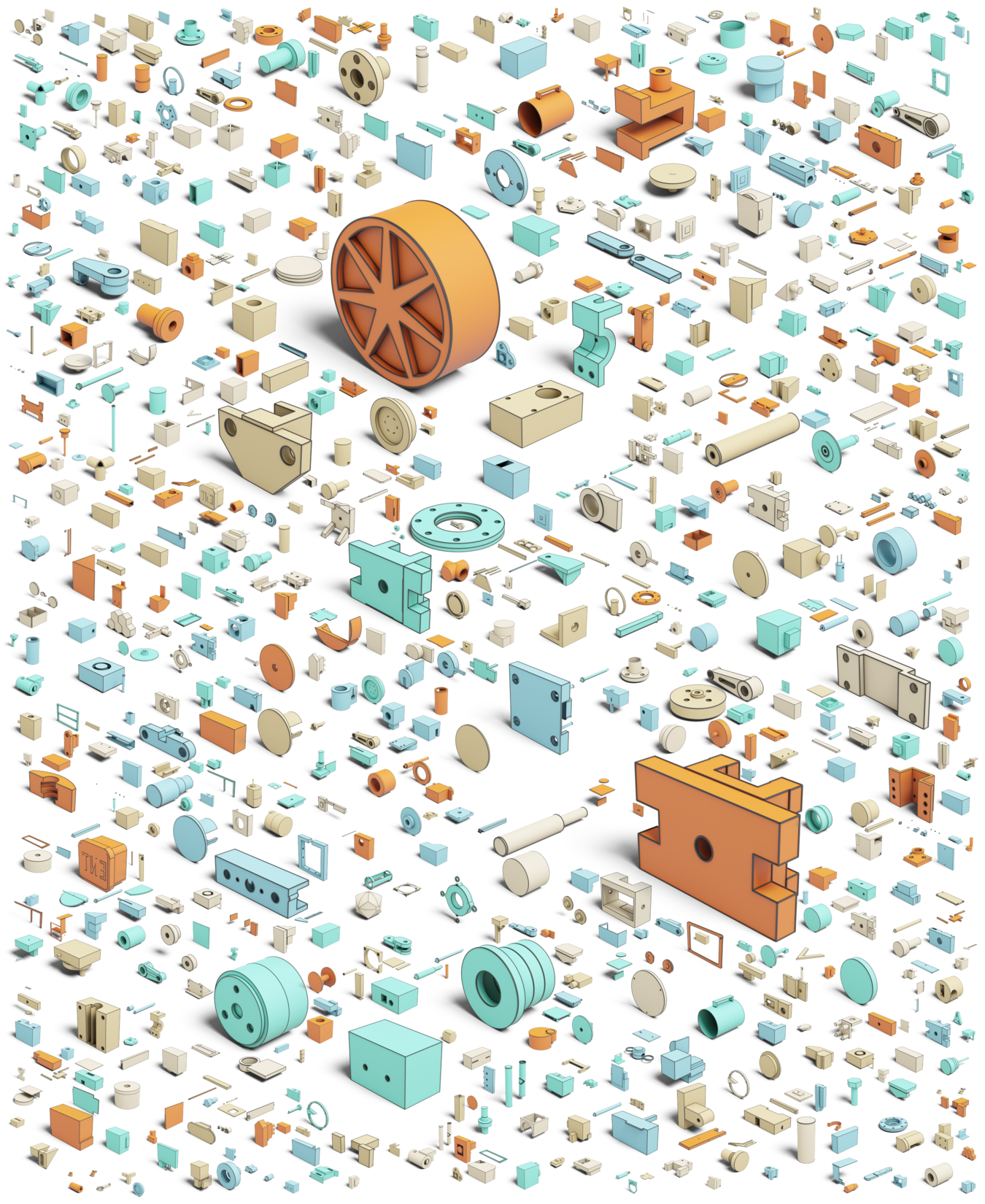}
    \caption{Randomly generated CAD models with three or more sketch-extrude steps from our method. }
    \label{fig:complex_mosaic_appendix}
\end{figure*}

\autoref{fig:cad_autocomplete} only shows the partial input and the final autocompleted CAD model. Here, we provide more examples in \autoref{fig:cad_ac_detailed} with detailed steps of the sketch-and-extrude from left to right. For every partial input, we show two generation of sketch-and-extrude CAD construction sequences due to the prediction of different code tree. \autoref{fig:mosaic_appendix} and \autoref{fig:complex_mosaic_appendix} provide additional randomly generated CAD models from our method.

\section{Sketch Generation Results}
\label{sec:appendix_sketch_results}

This section provides additional 2D sketch generation results conditioned on loop and profile codes. For random generation evaluation, we report the \textit{Fr\'{e}chet inception distance} (FID) score ~\cite{heusel2017gans} which computes the difference in mean and covariance for real and generated 2D data in a network feature space. Following SkexGen, we use ResNet-18~\cite{he2016deep} pre-trained on human sketch classification~\cite{eitz2012hdhso} for extracting the features to compute the FID.

\begin{figure*}
    \centering
    \includegraphics[width=0.9\textwidth]{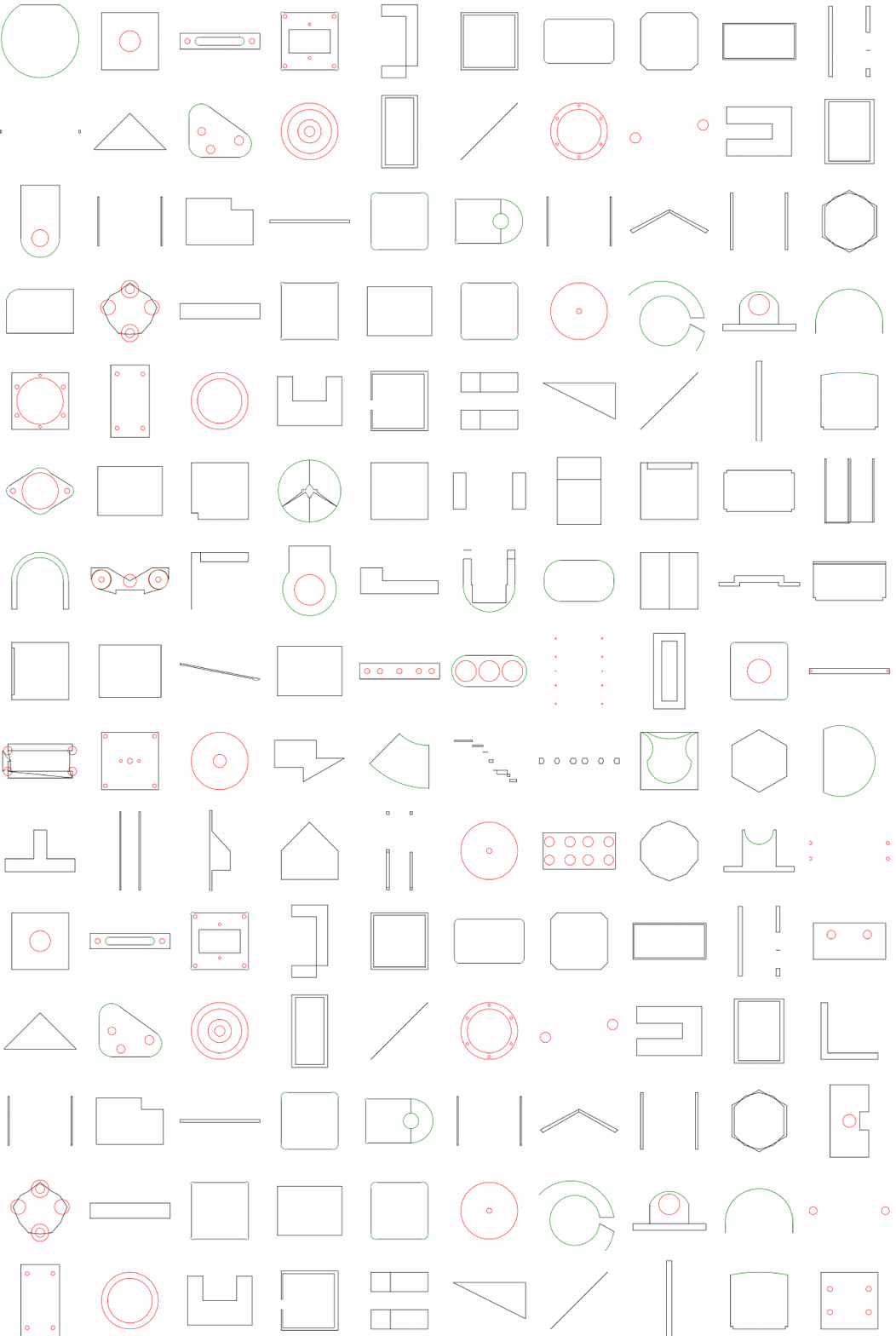}
    \caption{Randomly generated sketches from our method. }
    \label{fig:random_sketch5}
\end{figure*}

Quantitative numbers for our random sketch generation are reported in  
\autoref{tab:sketch_uncond}. Compared to DeepCAD and SkexGen, our model achieves a similar FID score as SkexGen but has slightly better \textit{Novel} score.  Qualitative result in \autoref{fig:random_sketch5} contains samples of randomly generated sketches from our method. Qualitatively, sketches from our method have complex shapes with few self-intersections or broken curves. 

\begin{figure*}
    \centering
    \includegraphics[width=0.99\textwidth]{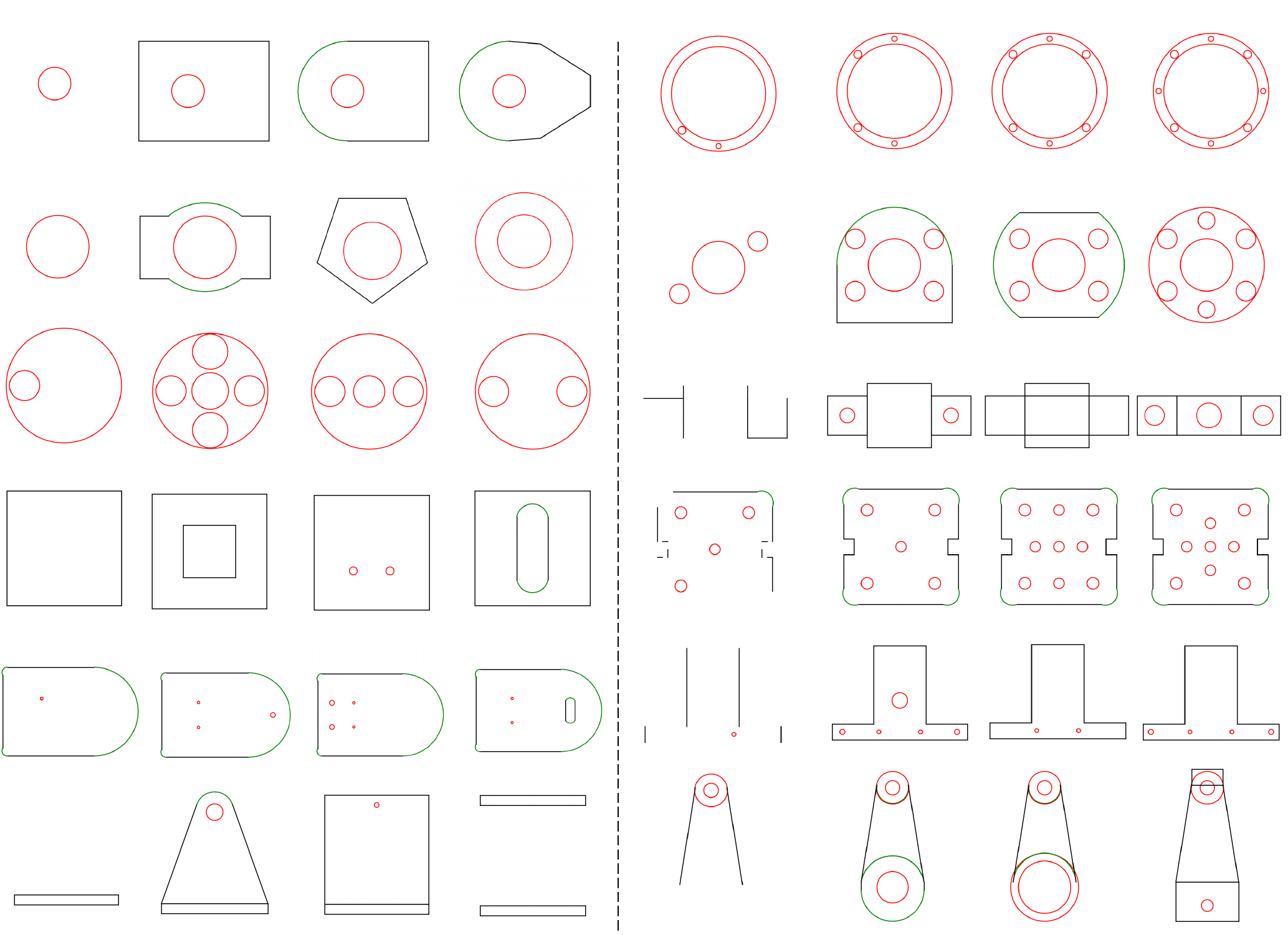}
    \caption{Left: Autocompleted sketches (column 2 $\sim$ 4) from partial loops (column 1). Right: Autocompleted sketches (column 6 $\sim$ 8) from partial curves (column 5).}
    \label{fig:appendix_sketch_ac1}
\end{figure*}

We provide additional sketch autocompletion results in \autoref{fig:appendix_sketch_ac1}. We also trained our model to autocomplete the full sketch from partial curves, which is a more general case for user interaction. 

\begin{table}[H]
\caption{Quantitative metrics for unconditional sketch generation with models trained on sketches from the DeepCAD dataset.}
\label{tab:sketch_uncond}
\begin{center}
\setlength\tabcolsep{3.5 pt}
\small
\begin{tabular}{lccc}
\toprule
Method & FID ($\downarrow$) & Unique (\%,$\uparrow$) & Novel (\%,$\uparrow$) \\ 
\midrule                 
DeepCAD          &   75.47         &  98.79 & 97.45 \\
SkexGen          &   18.56         &  96.02 & 83.54 \\
Ours             &   18.14         &  97.11 & 85.62  \\
\bottomrule
\end{tabular}
\end{center}
\end{table}

\section{Additional Code and Data Mapping}
\label{sec:appendix_code_code_mapping}
\begin{figure*}
    \centering
    \includegraphics[width=0.66\textwidth]{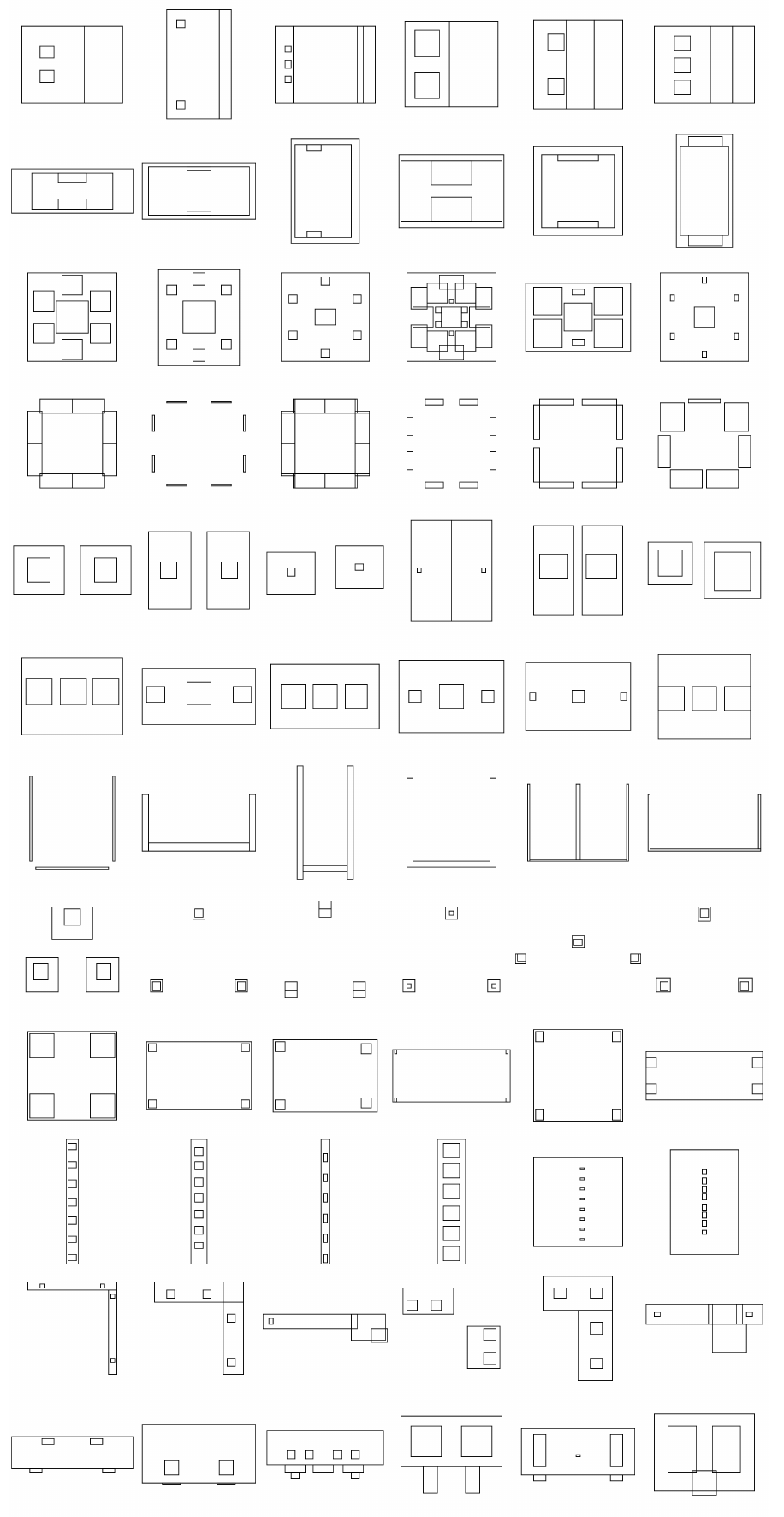}
    \caption{Profile data at each row are mapped to the same code.}
    \label{fig:cluster1}
\end{figure*}

\begin{figure*}
    \centering
    \includegraphics[width=0.66\textwidth]{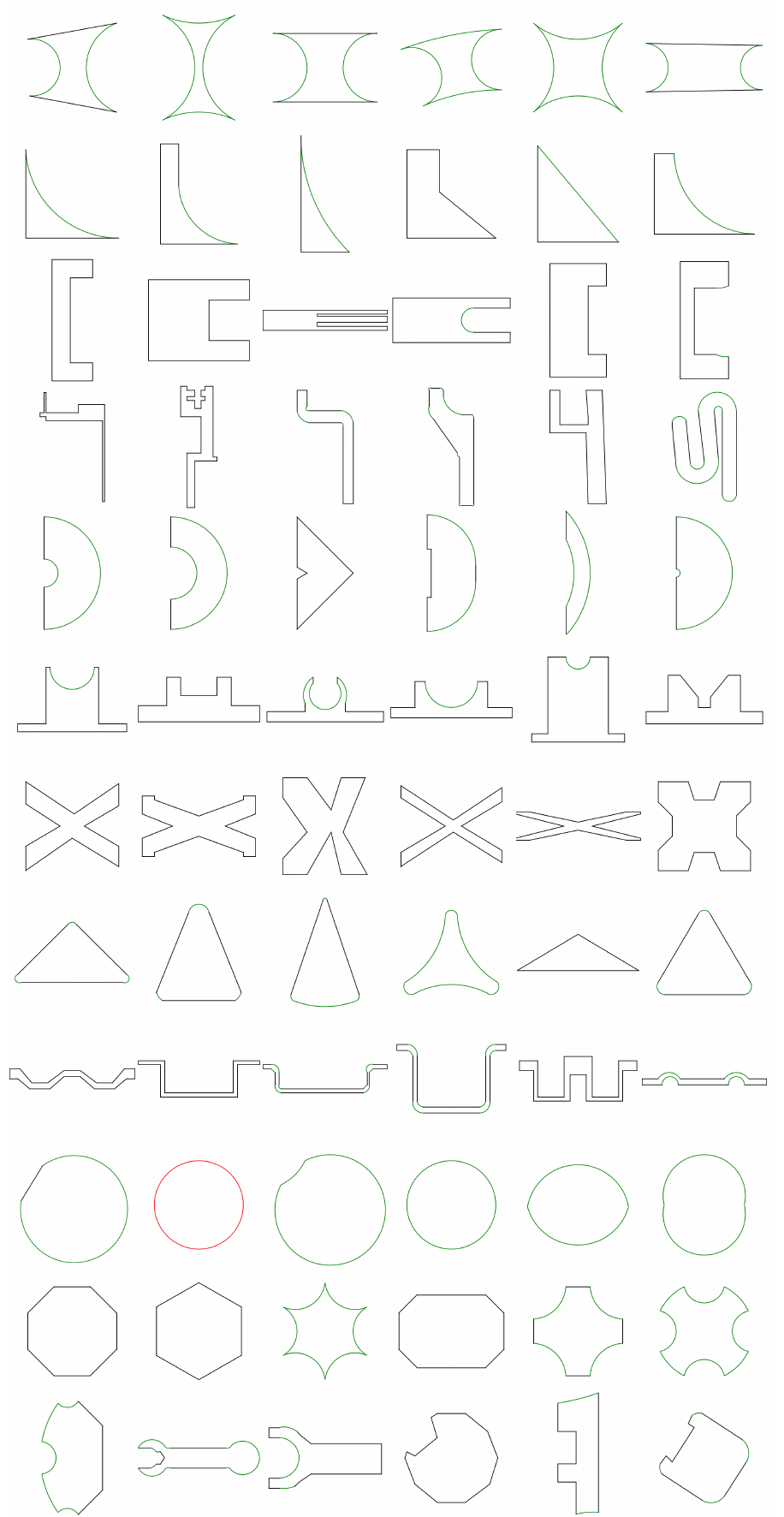}
    \caption{Loop data at each row are mapped to the same code.}
    \label{fig:cluster2}
\end{figure*}

Qualitative results from \autoref{fig:cluster1} and \autoref{fig:cluster2} contain more visualization of profile and loop data encoded to the same code. We see that data assigned to the same code also have similar design patterns, such as the circular layout around a center or the shape geometry approximating an 'X' shape. Those patterns ignore instance-specific details like the number of the bounding boxes, the number of curves, and the type of the curves.

\end{document}